\documentclass{article} % For LaTeX2e
\usepackage{iclr2023_conference,times}

\usepackage{amsmath,amsfonts,bm}

\def\eqref#1{equation~\ref{#1}}
\def\1{\bm{1}}

\DeclareMathAlphabet{\mathsfit}{\encodingdefault}{\sfdefault}{m}{sl}
\SetMathAlphabet{\mathsfit}{bold}{\encodingdefault}{\sfdefault}{bx}{n}

\usepackage[pagebackref,citecolor=blue,linkcolor=blue,colorlinks,urlcolor=blue,bookmarks=false]{hyperref}
\usepackage{url}
\usepackage{graphicx}
\usepackage{amsmath}
\usepackage{amssymb}
\usepackage{booktabs}
\usepackage{comment}

\usepackage{times}
\usepackage{epsfig}
\usepackage{subfig}
\usepackage{tabularx,booktabs,multirow,makecell}
\usepackage{caption}
\newcommand{\specialcell}[2][c]{\begin{tabular}[#1]{@{}c@{}}#2\end{tabular}}

\iclrfinalcopy

\title{Exploiting Category Names for Few-Shot Classification with Vision-Language Models}

\author{Taihong Xiao$^{1}$, Zirui Wang$^3$, Liangliang Cao$^{3}$, Jiahui Yu$^2$, Shengyang Dai$^2$, Ming-Hsuan Yang$^{1,2}$\\
{\{txiao3, mhyang\}@ucmerced.edu \quad \{jiahuiyu, sydai\}@google.com \quad \{ziruiw, llcao\}@apple.com} \\
$^1$ University of California, Merced \qquad $^2$ Google \qquad $^3$ Apple
}

\begin{document}

\maketitle

\begin{abstract}

Vision-language foundation models pretrained on large-scale data provide a powerful tool for many visual understanding tasks.
Notably, many vision-language models build two encoders (visual and textual) that can map two modalities into the same embedding space. As a result, the learned representations achieve good zero-shot performance on tasks like image classification. However, when there are only a few examples per category, the potential of large vision-language models is often underperformed, mainly due to the gap between a large number of parameters and a relatively small amount of training data. 
This paper shows that we can significantly improve the performance of few-shot classification by using the category names to initialize the classification head.  
With the proposed category name initialization method, our model obtains the state-of-the-art performance on a number of few-shot image classification benchmarks (e.g., 87.37\% on ImageNet and 96.08\% on Stanford Cars, both using five-shot learning). 

\end{abstract}

\section{Introduction}

In recent years, large vision-language models have opened doors to many new applications and provided new thoughts to existing problems. The advantages of large vision-language models are blessed by learning from largely available images with surrounding texts, as well as exploring the capacity of transformer network~\citep{dosovitskiy2021image} to model web-scale image-text data. \citeauthor{Radford2021LearningTV} first proposed CLIP for vision-language modeling, which was followed by numerous works, including ALIGN~\citep{jia2021scaling}, LiT~\citep{zhai2022lit}, Flamingo~\citep{Alayrac2022FlamingoAV}, Florence~\citep{Yuan2021FlorenceAN}, CoCa~\citep{Yu2022CoCaCC}, etc. The development of vision-language models provides novel perspectives of thinking of few-example learning. 

This paper considers the problem of few-shot classification in the new light of large vision-language models. 
Researchers have found that models pretrained from ImageNet can be easily transferred by finetuning on a new classification task \citep{huh2016makes}. Similarly, we can take the vision encoder from the pretrained vision-language model and finetune it with a few examples. Since state-of-the-art vision-language models were pretrained on billions of web images and texts, such finetuning often outperforms the models trained on ImageNet with better robustness and generalization capabilities. 

Despite the capability of the text branch in pretrained vision-language models, it is not optimally utilized when directly fine-tuning the vision component for downstream image classification tasks. Additionally, the large size of these models can lead to overfitting when trained on limited data.
In addition to the above approach, we exploit another source of information in vision-language models that traditional models have overlooked. Such new information comes from the category names in downstream image classification tasks. Because vision-language models can generate powerful representations for both images and texts, we will show that by utilizing semantic category names, vision-language models can be transferred better with few examples in downstream tasks.

\begin{figure}[htp]
    \centering
    \includegraphics[width=0.3\textwidth]{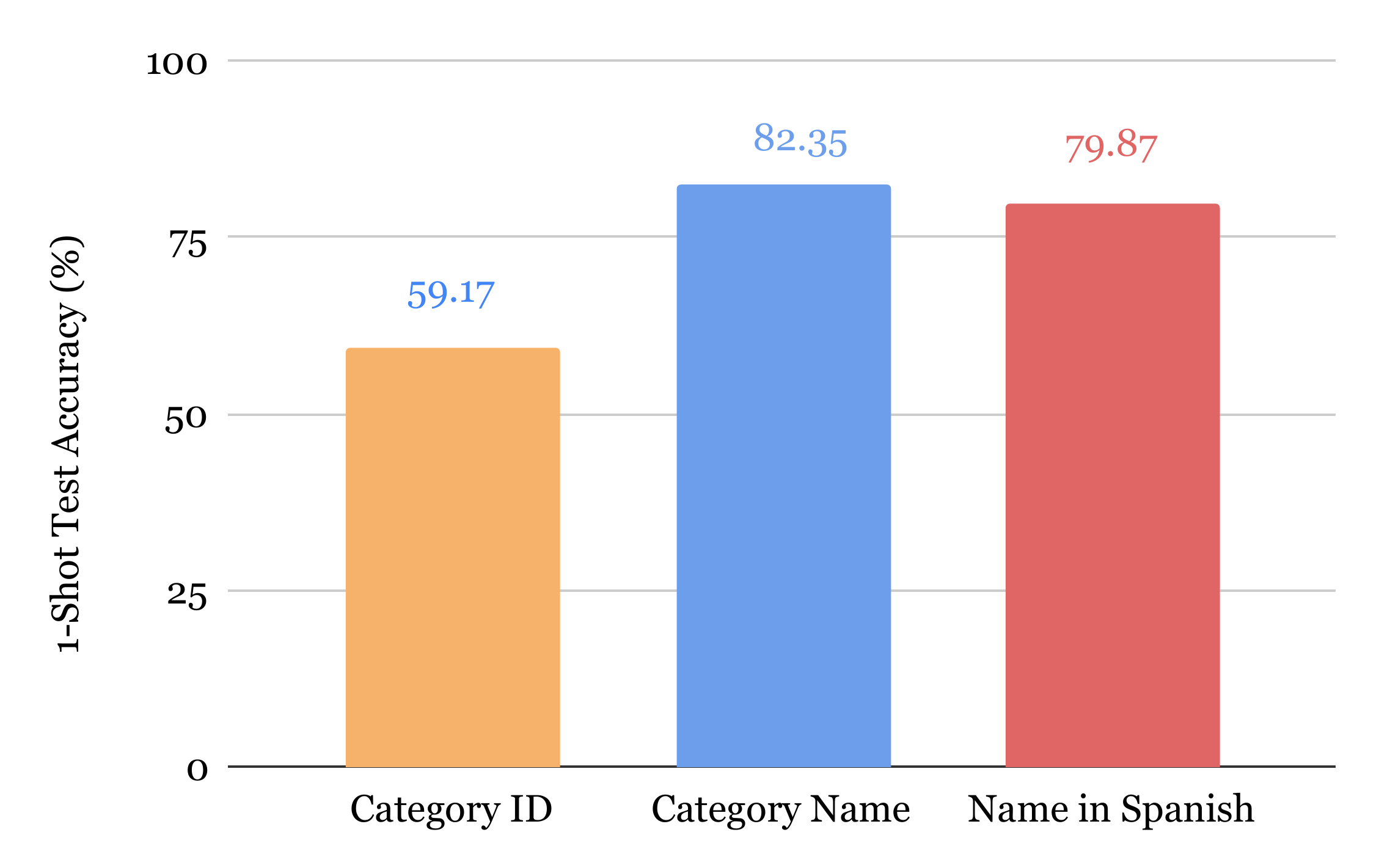}
    \hfil
    \includegraphics[width=0.55\textwidth]{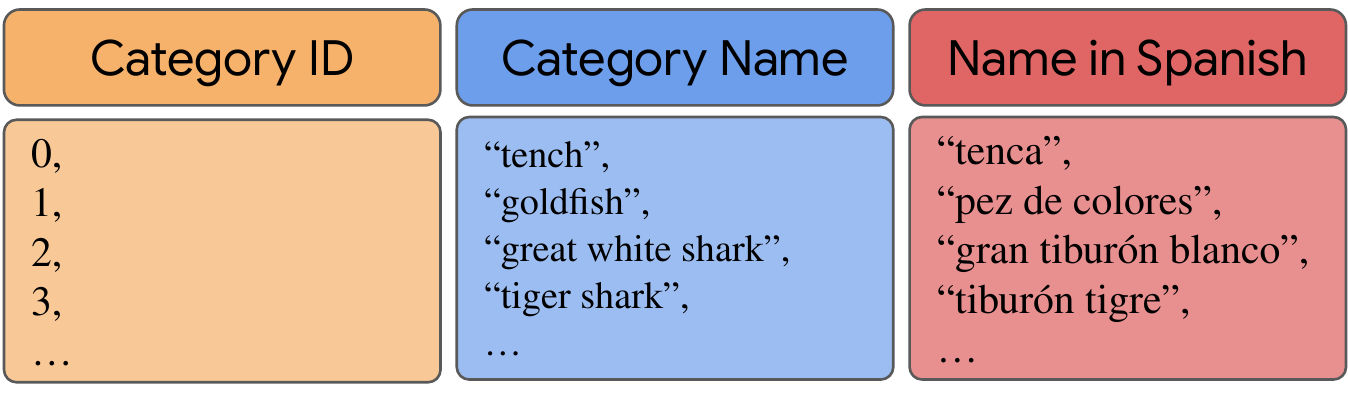}
    \caption{
    Comparing one-shot classification accuracy on ImageNet using different category information.
    % The typical way of finetuning using images with their category IDs does not work well for one-shot learning with big models. With the information on the category names of training images, we develop a new initialization approach that significantly boosts the performance of vision-language models in few-shot learning. Interestingly, using non-English names can still help even though the model was pre-trained using images and English text data pairs.
    }
    \label{fig:different_init}
    \vspace{-2mm}
\end{figure}

As summarized in Figure~\ref{fig:different_init}, this paper explores several scenarios: (1) randomly initializing a classification head; (2) initializing a classification head with category names; (3) initializing a classification head with other heuristics such as class digits or even non-English category names.
Note that (1) corresponds to the scenario when we only know the category ID (e.g., class 0, class 1, ..., class N) without knowing the meaning of each category. However, (2) implicitly parses the information from category names such as ``tench" and ``goldfish". These label names could be processed by the pretrained language model to provide a better initialization for the model adaption. 
As a comparison to (2), (3) provides different types of category name information.
The main difference between scenario (1) and the others is that (1) does not utilize text/language information from the categories. In scenario (1), the backbone network is initialized from the pretrained model weights, and the classification head is randomly initialized. We set (1) to be our baseline as it is the most common model adaptation method. For the other scenarios, we leverage the pretrained language model to parse the text information in the provided categories. Specifically, we pair all category names with prompts and extract the average text embedding as the weight to initialize the classification head. The second scenario is called {\it category name initialization} (CNI), and it has achieved the best performance among all these scenarios when finetuning using one-shot ImageNet data, as shown in Figure~\ref{fig:different_init}.

\section{Approach}

% In this section, we first briefly review CoCa~\citep{Yu2022CoCaCC}, one of the state-of-the-art vision-language models, and then discuss two initialization strategies: the standard random initialization and new category name initialization (CNI) for finetuning tasks. 

% \subsection{Revisiting CoCa pretraining}

\begin{figure}[!htb]
    \centering
    \includegraphics[width=0.99\textwidth]{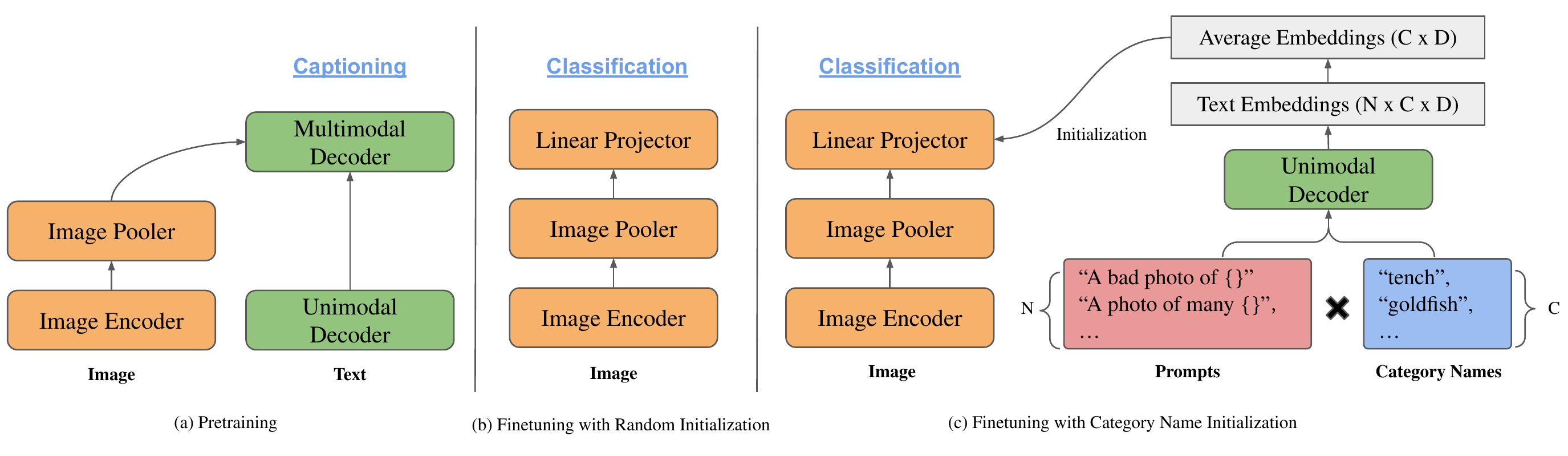}
    \caption{An overview of the CoCa pretraining and finetuning.
    (a) The pretraining of CoCa relies on mapping image and text pairs into the same space for embedding alignment, where the image and text embeddings are extracted through an image encoder and a unimodal text decoder, respectively. The image pooler is used to customize the image embedding for different tasks. (b) We append a randomly initialized linear projector to the image pooler and initialize the image encoder from pretrained weights. (c) We construct text sequences by pairing all $C$ category names with $N$ different prompts. Via the pretrained unimodal decoder, we can compute the text embeddings for all text sequences (with a total number of $N\times C$), each of them is $D$-dimensional vector. The normalized average embeddings can be used for initializing the weight in the linear projector.    
    }
    \label{fig:framework}
\end{figure}

In this paper, we take a recent state-of-the-art vision language model -- CoCa to illustrate our approach.
Unlike other recent vision-language models, CoCa adopts an encoder-decoder model architecture to learn generic vision and multimodal representations. As shown in Figure~\ref{fig:framework} (a), CoCa encodes images to latent representation via an encoder network (e.g., vision transformer (ViT)~\citep{dosovitskiy2021image}). An image pooler is appended after the image encoder to customize the image representations for different tasks and training objectives. 
On the other hand, CoCa uses a unimodal decoder to extract text-only embeddings and cascades multimodal decoder layers cross-attending to image embeddings to learn multimodal image-text representations. 
% 
% CoCa is pretrained on image-text pairs using two objective functions. The first is contrastive loss, where the image representations are contrasted against the paired text representations. The contrastive loss enables cross-modal representation alignment. The other is image captioning loss, which requires the model to autoregressively predict the tokenized texts by maximizing the conditional likelihood. 
% %
% The resulting CoCa can thus generate both unimodal visual/textual embeddings and multimodal joint embeddings.
% The unimodal visual output generated by the encoder and the unimodal textual output generated by the unimodal decoder are aligned in the same vector space and thus can be used to map images with their class names in a zero-shot manner.
%
Here, we focus on reusing these two components to initialize for few-shot learning.

% \subsection{Finetuning CoCa}\label{sec-approach-finetuning}

\paragraph{Random initialization.}
One straightforward model adaption approach is to add a randomly initialized linear projector upon the pretrained model and selectively finetune the model (all or part of the layers), as depicted in Figure~\ref{fig:framework} (b). 
Following the approach used by CLIP~\citep{Radford2021LearningTV} and CoCa~\citep{Yu2022CoCaCC}, we first use an image pooler to obtain the aggregated image embedding $H\in\mathcal{R}^{D}$ and then apply a linear projector to get the prediction $Y\in\mathcal{R}^{C}$,
\begin{align}
Y = \mathrm{softmax}(WH + b),
\end{align}
where $W\in\mathcal{R}^{C\times D}$ and $b\in\mathcal{R}^{C}$ are learnable weight and bias of the linear projector. Here $W$ and $b$ are randomly initialized, while the image encoder and generative image pooler are initialized from the pretrained weights.

\paragraph{Category name initialization.}
We argue that the above random initialization ignores the potential of the language model for model adaptation. In contrast, we propose the category name initialization to maximize the capacity of the pretrained unimodal decoder.
First, we pair all category names (whose total number is $C$) with $N$ different prompts as the text inputs. For example, pairing the category name ``tench'' with a prompt ``A bad photo of \{\}'' gives us a text sequence ``A bad photo of tench''. 
Next, we compute the text embeddings for all these $N\times C$ text sequences via the unimodal decoder. As the text embedding for each text input is a $D$-dimensional vector, we can obtain a text embedding tensor with a shape of $N\times C\times D$. 
We then compute the average over different prompts and perform the normalization to obtain the average embeddings of shape $C\times D$. 
Unlike random initialization, we initialize the weight $W$ by the average embeddings and bias $b$ by a zero vector in the linear projector. 
To enable zero-shot inference of the category name initialized model, we use the pretrained model weights to initialize the image encoder and the image pooler.

\section{Experiments}

We conduct our finetuning experiments on several widely-used image classification datasets, including ImageNet~\citep{Deng2009ImageNetAL}, ImageNet-V2~\citep{Recht2019DoIC}, ImageNet-R~\citep{Hendrycks2021TheMF}, ImageNet-A~\citep{Hendrycks2021NaturalAE},  ImageNet-Sketch~\citep{Wang2019LearningRG}, Cifar100~\citep{krizhevsky2009learning}, Oxford Flowers~\citep{Nilsback2008AutomatedFC}, Stanford Cars~\citep{Krause20133DOR}, Country-211~\citep{Radford2021LearningTV}, Food-101~\citep{Bossard2014Food101M}, FGVC Aircraft~\citep{Maji2013FineGrainedVC}, EuroSAT~\citep{Helber2019EuroSATAN}, and Oxford-IIIT Pets~\citep{parkhi2012cats}. For different few-shot settings, we randomly sample a particular portion of data from these datasets. For example, one-shot ImageNet means that we only select one image from the ImageNet training data for each category. The evaluation is still performed using the whole testing set. Following the existing benchmark~\citep{Li2022ELEVATERAB}, we use the same text prompts\footnote{\url{https://github.com/Computer-Vision-in-the-Wild/Elevater_Toolkit_IC/blob/main/vision_benchmark/datasets/prompts.py}} for evaluating all methods.

We use the pretrained CoCa model and apply category name initialization. Following the standard benchmark evaluation~\citep{Li2022ELEVATERAB}, we then compare our method against the previous works, including 
MAE~\citep{He2022MaskedAA}, CLIP~\citep{Radford2021LearningTV}, CLIP+CoOp~\citep{Zhou2022LearningTP}, WiSE-FT~\citep{wortsman2022robust}, and Flamingo-3B~\citep{Alayrac2022FlamingoAV} on ImageNet and its variants, including ImageNet-V2~\citep{Recht2019DoIC}, ImageNet-R~\citep{Hendrycks2021TheMF}, ImageNet-A~\citep{Hendrycks2021NaturalAE} and ImageNet-Sketch~\citep{Wang2019LearningRG}. As shown in Table~\ref{tab:few-shot-imagenet-variants}, the CoCa-2B model has achieved state-of-the-art few-shot classification results on all these benchmarks. Surprisingly, the one-shot performance of CoCa-base is even better than the performance of some recent methods finetuned on the whole dataset.

In addition to ImageNet and variants, we show that our method can achieve state-of-the-art few-shot performance on other image classification benchmarks, including  Cifar100~\citep{krizhevsky2009learning}, Oxford Flowers~\citep{Nilsback2008AutomatedFC} and Stanford Cars~\citep{Krause20133DOR}, Country-211~\citep{Radford2021LearningTV}, Food-101~\citep{Bossard2014Food101M}, FGVC Aircraft~\citep{Maji2013FineGrainedVC}, EuroSAT~\citep{Helber2019EuroSATAN}, and Oxford-IIIT Pets~\citep{parkhi2012cats}. 
As a comparison, we choose MAE~\citep{He2022MaskedAA}, CAE~\citep{ContextAutoencoder2022}, MoCo-v3~\citep{Chen2021AnES}, DeiT~\citep{Touvron2021TrainingDI}, ViT~\citep{dosovitskiy2021image} and CLIP~\citep{Radford2021LearningTV}. 
We can observe from Table~\ref{tab:few-shot-benchmark} that our CoCa-2B model outperforms many other approaches even with less training data.
Such good performance benefits a lot from the category name initialization, as it provides a good starting point so that the model could achieve better performance using a few examples.

\begin{table}[tp]
  \centering
  \renewcommand*{\arraystretch}{1.2}
  \caption{Few-shot results on ImageNet and its variants. We use IN as the abbreviation for ImageNet, and CNI for category name initialization. The second column means how much training data per class is used for finetuning. 0 shot means the pretrained vision-language model is directly evaluated without finetuning. Full means the entire training set has been used. All the numbers under the last five columns denote the top-1 test accuracy. 
  }
  \label{tab:few-shot-imagenet-variants}
  \scalebox{0.8}{
    \begin{tabular}{rcccccc}
        \Xhline{1pt}
         Model & Shot & IN & IN-V2  & IN-R & IN-A & IN-Sketch \\
        \hline
        MAE & full & - & - & 66.50 & 76.70 & 50.90  \\
        \hline
        \multirow{2}[0]{*}{CLIP (ViT-B/16)} & 0 & 68.40 & 62.60 & 77.60 & 50.00 & 48.20 \\
         & full &  79.90 & 69.80 &  70.80 & 46.40 &  46.90 \\
        \hline
        \multirow{2}[0]{*}{CLIP (ViT-L/14)}  & 0     & 76.20 & 70.10 & 88.90 & 77.2 & 60.20 \\
          & full  & 85.20 & 75.80 & 85.30 & 76.10 & 58.70 \\
        \hline
        \multirow{3}[0]{*}{CLIP+Adapter (ResNet-50)} & 0 & 55.50 & - & - & - & - \\
        & 1 & 58.10 & - & - & - & - \\
        & 4 & 59.50 & - & - & - & - \\
        \hline
        \multirow{2}[0]{*}{CLIP+CoOp (ViT-B/16)} & 0 & 58.18 & -& - & - & - \\
         & 1 & 58.00 & - & - & - & - \\
         & 4 & 60.01 & -  &  - & - & - \\
        \hline
        \multirow{3}[0]{*}{Tip-Adapter-F (ResNet-50)} & 0 & 60.33 & - & - & - & - \\
         & 1 & 61.32 & - & - & - & - \\
         & 4 & 62.52 & - & - & - & - \\
        \hline
        WiSE-FT (ViT-L/14) & full  & 85.30 & 76.90 & 89.80 & 79.70 & 63.00 \\
        \hline
        \multirow{2}[0]{*}{Flamingo-3B} & 1 &  70.90 & - & - & - & - \\
         & 5 &  72.70 & - & - & - & - \\
        \hline
        \multirow{2}[0]{*}{Flamingo-80B} & 1 & 71.90 & - & - & - & - \\
         & 5 & 77.30 & - & - & - & -  \\
        \hline
        CoCa-base & 0 & 82.26 & 76.22 & 93.16 & 76.17 & 71.12 \\
        \hline
        % \multirow{3}[0]{*}{CoCa-base (Ours)} & 0 & 82.26 & 76.62 & 93.16 &  76.17  & \\
         \multirow{2}[0]{*}{CoCa-base+CNI (Ours)} & 1 & 82.35 & 76.47 & 93.37 & 77.00 & 71.61 \\
         & 5 & 83.58 &  77.23 & 93.22  &  77.23 &  71.35 \\
          % & 1\% ($\approx12.8$) & 83.91 & 75.57 & 92.21 & 71.95 & 69.73 \\
        \hline
        CoCa-2B & 0 & 86.09 & 80.39 & 96.19 & 89.39 & 77.12 \\
        \hline
        % \multirow{3}[0]{*}{CoCa-2B (Ours)} & 0 & 86.19 & 80.39 & 96.49 & 89.99 & 77.52 \\
        \multirow{2}[0]{*}{CoCa-2B+CNI (Ours)
        } & 1 & 86.15 & 80.57 & 96.62 & 90.12 & 77.49 \\
        & 5 & 87.37 & 81.66 & 96.41 & 89.68 & 77.39 \\
         % & 1\% & 87.90 & 81.75 & 96.25 & 90.19 & 76.34 \\
        \Xhline{1pt}
    \end{tabular}%
  }
\end{table}

\begin{table}[tp]
  \centering
  \renewcommand*{\arraystretch}{1.3}
  \caption{Comparing with the state-of-the-art on multiple classification benchmarks. CNI stands for category name initialization. Our model obtains the state-of-the-art few-shot learning performance with less training data than many others. }
  \label{tab:few-shot-benchmark}
  \scalebox{.78}{
    \begin{tabular}{rccccccccc}
        \Xhline{1pt}
         Model & Shot & Cifar100   & \specialcell{Oxford\\ Flowers} & \specialcell{Stanford\\ Cars} & Country-211 & Food-101 & \specialcell{FGVC\\ Aircraft} & EuroSAT & \specialcell{Oxford-IIIT\\ Pets} \\
        \hline
        \multirow{3}[0]{*}{MAE} & 5 &  21.20 &  50.90 & 6.30 & \phantom{0}2.80 &  \phantom{0}7.70 & \phantom{0}7.00 & 64.60 & 17.20 \\
         & 20 & 43.50  & 71.90 & 25.50 & \phantom{0}4.40  & 30.40 & 29.90 & 74.10 & 60.00 \\
         & full & 68.30 & 72.00 & 37.20 & 10.10 & 65.10 & 39.10 & 94.80 & 81.60  \\
        \hline
        \multirow{3}[0]{*}{CAE} & 5 & 38.30  & 70.30 & 8.70 & \phantom{0}3.50 & 18.60 & 14.30 & 76.70 & 37.30\\
         & 20 & 55.10  & 81.20 & 27.50 & \phantom{0}5.50   & 35.70 & 32.60 & 89.00 & 63.30  \\
         & full & 78.90 & 81.20 & 40.40 & 11.40 & 67.40 & 40.80 & 96.70 & 79.80  \\
        \hline
        \multirow{3}[0]{*}{MoCo-v3} & 5  & 60.50 & 79.50 & 13.40 & \phantom{0}4.80 & 36.60 & 11.80 & 77.10 & 76.20 \\
         & 20 & 75.50 &  89.50 & 49.50 & \phantom{0}7.60 &  59.30 & 38.20 &  84.80 & 86.40 \\
         & full & 85.30 & 89.50 & 63.00 & 13.70 & 78.00 & 48.00 & 95.90 & 91.40 \\
        \hline
        \multirow{3}[0]{*}{DeiT} & 5  & 61.50 & 82.70 & 27.60 & \phantom{0}4.40 & 41.90 & 24.10 & 62.50 & 87.80  \\
          & 20 & 73.70 & 92.70 & 68.80 & \phantom{0}6.20 & 61.50 & 34.10 & 90.70 & 91.90 \\
          & full & 89.60 &   92.40 & 83.00 & 14.10 &  84.50 & 59.30 & 98.20 & 93.90 \\
        \hline
        \multirow{3}[0]{*}{ViT} & 5  & 75.40  &  99.20 & 27.60 & \phantom{0}6.80  &  59.00 & 22.70 & 70.00 & 89.60\\
         & 20 &  84.00 &  99.20 & 53.90 & 11.50 &  81.70 & 40.50 &  86.50 &  92.60 \\
         & full & 89.80 &  99.20 & 67.50 & 16.60 & 89.60 & 47.80 & 96.00 &  94.80 \\
        \hline
        \multirow{2}[0]{*}{CLIP} & 5 & 71.10  & 94.20 & 73.60 & 21.70  & 89.70 & 36.00 & 76.70 & 90.50 \\
             & 20  & 75.40 &  96.8 & 73.60  & 25.20 &  90.60 & 48.10 & 86.60 & 92.30 \\
        \hline 
        CoCa-2B & 0 & 77.19 & 92.04 & 94.37  & 42.15 & 94.79 & 44.83 & 49.74 & 97.88\\
        \hline
        \multirow{3}[0]{*}{CoCa-2B+CNI} 
        % & 0 & 77.19 & 92.04 & 94.37  & 42.15 & 94.79 & 44.83 & 49.74 & 97.88\\
          & 1  & 77.89 & 98.45 & 95.29 & 42.44 & 94.91 & 58.33 & 75.06 & 97.93  \\
          & 5  & 78.62 & 99.25 & 96.08 & 44.52 & 95.50 & 69.29 & 85.78 & 98.12 \\
        \Xhline{1pt}
    \end{tabular}%
  }
\end{table}%

\section{Related Work}

Recently, there has been increasing interest in utilizing the vision-language model for visual zero-shot learning, a related problem of few-shot learning.
CLIP~\citep{Radford2021LearningTV} is a pioneering work in large-scale vision-language modeling. 
Unlike previous works in vision-language representation~\citep{donahue2015long,vinyals2015show},  CLIP collects image-text pairs from the Web, which contains diversified semantics in a weakly supervised fashion. In addition, CLIP is built on large-scale contrastive learning, which maps images and text into the same subspace. 
Through this, the model can map textual class names with images hence performing image classification in a zero-shot manner.
The approach of CLIP was followed by ALIGN~\citep{jia2021scaling},
Flamingo~\citep{Alayrac2022FlamingoAV}, LiT~\citep{zhai2022lit}, Florence~\citep{Yuan2021FlorenceAN}, FLAVA~\citep{singh2022flava}, SimVLM~\citep{DBLP:conf/iclr/WangYYDT022} and CoCa~\citep{Yu2022CoCaCC}. Among these works, ALIGN, Florence, FLAVA, and LIT are based on contrastive learning. Flamingo chooses to optimize a generative loss with gated cross-attention layers. At last, CoCa integrates contrastive and generative loss into one framework. Although training CoCa seems the most challenging among all these vision-language works, it obtains consistently better results in many tasks.

In the literature, CLIP, LiT, ALIGN, Florence, FLAVA, and CoCa have demonstrated promising results with zero-shot learning.  
However, the potential of these models for few-shot learning is not well exploited.  
\cite{Li2022ELEVATERAB} construct a benchmark and toolkit named Elevater for evaluating the transferability of vision-language models using different training samples. 
\cite{Radford2021LearningTV} point out that using few training examples could improve the effectiveness robustness while undermining the relative robustness.
Most few-shot learning algorithms are trained exclusively on image data, which ignores the valuable text information that can be used to enhance the learning process. However, Flamingo has emerged as a promising approach for addressing this issue. Flamingo utilizes few-shot interleaved prompts that incorporate gated cross-attention layers to improve few-shot learning.

\cite{Alayrac2022FlamingoAV} propose context optimization (CoOp) as a means of modeling text in prompts through continuous representations. \cite{zhou2022cocoop} propose CoCoOp, which extends CoOp by further learning a lightweight neural network to generate an input-conditional token (vector) for each image. In addition, there are a series of prior-based methods that utilize CLIP priors with a cache model. CLIP-Adapter~\citep{gao2021clip} combines zero-shot visual or language embeddings with corresponding finetuning features to improve performance. TIP-Adapter~\citep{zhang2021tip} constructs adapters using a key-value cache model from few-shot training sets and updates their prior knowledge through feature retrieval. TIP-X~\citep{Udandarao2022SuSXTN} further constructs an affinity matrix by measuring the KL divergence between test and few-shot samples, which removes direct reliance on the uncalibrated image-image similarities. APE~\citep{Zhu2023NotAF} explores the trilateral affinities between the test image, prior cache model, and textual representations and only enable a lightweight category-residual module to be trained. Among these approaches, TIP-Adapter, TIP-X, and APE are training-free, while CoOp, CoCoOp, CLIP-Adapter, and APE-T~\citep{Zhu2023NotAF} are training required.
%
% This paper demonstrates that leveraging category names for initialization can significantly enhance the few-shot performance of the CoCa model without bells and whistles. Our approach not only outperforms the previously reported performance of Flamingo and CLIP, but also establishes a new state-of-the-art for both ImageNet and several other datasets with fewer training examples.

\section{Conclusion}

This paper has studied the few-shot classification problem using large vision-language models. 
Since it is hard to optimize large vision-language models with a few training examples, we propose exploring category names to initialize the classification head, significantly improving performance. 
In addition, we have also investigated the condition when the category names help. We demonstrate that borrowing other non-perfect category names or even names from a foreign language could also help the few-shot classification of vision-language models, which is better than randomly initializing the classification head. 
% However, the contribution of category names diminishes when the number of training samples becomes large.   
This paper obtains state-of-the-art few-shot performance on numerous benchmarks, including ImageNet, ImageNet-V2,  ImageNet-R, ImageNet-A, ImageNet-Sketch, Cifar100, Oxford Flowers, Stanford Cars, Country-211, Food-101, FGVC Aircraft, EuroSAT, and Oxford-IIIT Pets. Our few-shot classification result is even better than many previous works that have employed the whole training set.

\clearpage

\bibliography{iclr2023_conference}

\begin{thebibliography}{40}
\providecommand{\natexlab}[1]{#1}
\providecommand{\url}[1]{\texttt{#1}}
\expandafter\ifx\csname urlstyle\endcsname\relax
  \providecommand{\doi}[1]{doi: #1}\else
  \providecommand{\doi}{doi: \begingroup \urlstyle{rm}\Url}\fi

\bibitem[Alayrac et~al.(2022)Alayrac, Donahue, Luc, Miech, Barr, Hasson, Lenc,
  Mensch, Millican, Reynolds, Ring, Rutherford, Cabi, Han, Gong, Samangooei,
  Monteiro, Menick, Borgeaud, Brock, Nematzadeh, Sharifzadeh, Binkowski,
  Barreira, Vinyals, Zisserman, and Simonyan]{Alayrac2022FlamingoAV}
Jean-Baptiste Alayrac, Jeff Donahue, Pauline Luc, Antoine Miech, Iain Barr,
  Yana Hasson, Karel Lenc, Arthur Mensch, Katie Millican, Malcolm Reynolds,
  Roman Ring, Eliza Rutherford, Serkan Cabi, Tengda Han, Zhitao Gong, Sina
  Samangooei, Marianne Monteiro, Jacob Menick, Sebastian Borgeaud, Andy Brock,
  Aida Nematzadeh, Sahand Sharifzadeh, Mikolaj Binkowski, Ricardo Barreira,
  Oriol Vinyals, Andrew Zisserman, and Karen Simonyan.
\newblock Flamingo: a visual language model for few-shot learning.
\newblock In \emph{{Neural Information Processing Systems (NeurIPS)}}, 2022.

\bibitem[Bossard et~al.(2014)Bossard, Guillaumin, and
  Gool]{Bossard2014Food101M}
Lukas Bossard, Matthieu Guillaumin, and Luc~Van Gool.
\newblock Food-101 - mining discriminative components with random forests.
\newblock In \emph{{European Conference on Computer Vision (ECCV)}}, 2014.

\bibitem[Chen et~al.(2022)Chen, Ding, Wang, Xin, Mo, Wang, Han, Luo, Zeng, and
  Wang]{ContextAutoencoder2022}
Xiaokang Chen, Mingyu Ding, Xiaodi Wang, Ying Xin, Shentong Mo, Yunhao Wang,
  Shumin Han, Ping Luo, Gang Zeng, and Jingdong Wang.
\newblock Context autoencoder for self-supervised representation learning.
\newblock \emph{ArXiv}, 2202.03026, 2022.

\bibitem[Chen et~al.(2021)Chen, Xie, and He]{Chen2021AnES}
Xinlei Chen, Saining Xie, and Kaiming He.
\newblock An empirical study of training self-supervised vision transformers.
\newblock In \emph{{IEEE International Conference on Computer Vision (ICCV)}},
  pp.\  9620--9629, 2021.

\bibitem[Cubuk et~al.(2020)Cubuk, Zoph, Shlens, and Le]{Cubuk2020RandaugmentPA}
Ekin~Dogus Cubuk, Barret Zoph, Jonathon Shlens, and Quoc~V. Le.
\newblock Randaugment: Practical automated data augmentation with a reduced
  search space.
\newblock \emph{{IEEE Conference on Computer Vision and Pattern Recognition
  Workshops (CVPRW)}}, pp.\  3008--3017, 2020.

\bibitem[Deng et~al.(2009)Deng, Dong, Socher, Li, Li, and
  Fei-Fei]{Deng2009ImageNetAL}
Jia Deng, Wei Dong, Richard Socher, Li-Jia Li, K.~Li, and Li~Fei-Fei.
\newblock Imagenet: A large-scale hierarchical image database.
\newblock In \emph{{IEEE Conference on Computer Vision and Pattern Recognition
  (CVPR)}}, pp.\  248--255, 2009.

\bibitem[Donahue et~al.(2015)Donahue, Anne~Hendricks, Guadarrama, Rohrbach,
  Venugopalan, Saenko, and Darrell]{donahue2015long}
Jeffrey Donahue, Lisa Anne~Hendricks, Sergio Guadarrama, Marcus Rohrbach,
  Subhashini Venugopalan, Kate Saenko, and Trevor Darrell.
\newblock Long-term recurrent convolutional networks for visual recognition and
  description.
\newblock In \emph{{IEEE Conference on Computer Vision and Pattern Recognition
  (CVPR)}}, pp.\  2625--2634, 2015.

\bibitem[Dosovitskiy et~al.(2021)Dosovitskiy, Beyer, Kolesnikov, Weissenborn,
  Zhai, Unterthiner, Dehghani, Minderer, Heigold, Gelly,
  et~al.]{dosovitskiy2021image}
Alexey Dosovitskiy, Lucas Beyer, Alexander Kolesnikov, Dirk Weissenborn,
  Xiaohua Zhai, Thomas Unterthiner, Mostafa Dehghani, Matthias Minderer, Georg
  Heigold, Sylvain Gelly, et~al.
\newblock An image is worth 16x16 words: Transformers for image recognition at
  scale.
\newblock In \emph{{International Conference on Learning Representations
  (ICLR)}}, 2021.

\bibitem[Gao et~al.(2021)Gao, Geng, Zhang, Ma, Fang, Zhang, Li, and
  Qiao]{gao2021clip}
Peng Gao, Shijie Geng, Renrui Zhang, Teli Ma, Rongyao Fang, Yongfeng Zhang,
  Hongsheng Li, and Yu~Qiao.
\newblock Clip-adapter: Better vision-language models with feature adapters.
\newblock \emph{ArXiv}, 2110.04544, 2021.

\bibitem[He et~al.(2022)He, Chen, Xie, Li, Doll'ar, and
  Girshick]{He2022MaskedAA}
Kaiming He, Xinlei Chen, Saining Xie, Yanghao Li, Piotr Doll'ar, and Ross~B.
  Girshick.
\newblock Masked autoencoders are scalable vision learners.
\newblock In \emph{{IEEE Conference on Computer Vision and Pattern Recognition
  (CVPR)}}, pp.\  15979--15988, 2022.

\bibitem[Helber et~al.(2019)Helber, Bischke, Dengel, and
  Borth]{Helber2019EuroSATAN}
Patrick Helber, Benjamin Bischke, Andreas~R. Dengel, and Damian Borth.
\newblock Eurosat: A novel dataset and deep learning benchmark for land use and
  land cover classification.
\newblock \emph{IEEE Journal of Selected Topics in Applied Earth Observations
  and Remote Sensing}, 12:\penalty0 2217--2226, 2019.

\bibitem[Hendrycks et~al.(2021{\natexlab{a}})Hendrycks, Basart, Mu, Kadavath,
  Wang, Dorundo, Desai, Zhu, Parajuli, Guo, Song, Steinhardt, and
  Gilmer]{Hendrycks2021TheMF}
Dan Hendrycks, Steven Basart, Norman Mu, Saurav Kadavath, Frank Wang, Evan
  Dorundo, Rahul Desai, Tyler~Lixuan Zhu, Samyak Parajuli, Mike Guo,
  Dawn~Xiaodong Song, Jacob Steinhardt, and Justin Gilmer.
\newblock The many faces of robustness: A critical analysis of
  out-of-distribution generalization.
\newblock In \emph{{IEEE International Conference on Computer Vision (ICCV)}},
  pp.\  8320--8329, 2021{\natexlab{a}}.

\bibitem[Hendrycks et~al.(2021{\natexlab{b}})Hendrycks, Zhao, Basart,
  Steinhardt, and Song]{Hendrycks2021NaturalAE}
Dan Hendrycks, Kevin Zhao, Steven Basart, Jacob Steinhardt, and Dawn~Xiaodong
  Song.
\newblock Natural adversarial examples.
\newblock In \emph{{IEEE Conference on Computer Vision and Pattern Recognition
  (CVPR)}}, pp.\  15257--15266, 2021{\natexlab{b}}.

\bibitem[Huh et~al.(2016)Huh, Agrawal, and Efros]{huh2016makes}
Minyoung Huh, Pulkit Agrawal, and Alexei~A Efros.
\newblock What makes imagenet good for transfer learning?
\newblock \emph{ArXiv}, abs/1608.08614, 2016.

\bibitem[Jia et~al.(2021)Jia, Yang, Xia, Chen, Parekh, Pham, Le, Sung, Li, and
  Duerig]{jia2021scaling}
Chao Jia, Yinfei Yang, Ye~Xia, Yi-Ting Chen, Zarana Parekh, Hieu Pham, Quoc Le,
  Yun-Hsuan Sung, Zhen Li, and Tom Duerig.
\newblock Scaling up visual and vision-language representation learning with
  noisy text supervision.
\newblock In \emph{{International Conference on Learning Representations
  (ICLR)}}, pp.\  4904--4916, 2021.

\bibitem[Krause et~al.(2013)Krause, Stark, Deng, and Fei-Fei]{Krause20133DOR}
Jonathan Krause, Michael Stark, Jia Deng, and Li~Fei-Fei.
\newblock 3d object representations for fine-grained categorization.
\newblock \emph{{IEEE International Conference on Computer Vision Workshops
  (ICCVW)}}, pp.\  554--561, 2013.

\bibitem[Krizhevsky(2009)]{krizhevsky2009learning}
Alex Krizhevsky.
\newblock Learning multiple layers of features from tiny images.
\newblock Technical report, University of Toronto, 2009.

\bibitem[Kudo \& Richardson(2018)Kudo and Richardson]{Kudo2018SentencePieceAS}
Taku Kudo and John Richardson.
\newblock Sentencepiece: A simple and language independent subword tokenizer
  and detokenizer for neural text processing.
\newblock In \emph{{Conference on Empirical Methods in Natural Language
  Processing (EMNLP)}}, 2018.

\bibitem[Li et~al.(2022)Li, Liu, Li, Zhang, Aneja, Yang, Jin, Lee, Hu, Liu, and
  Gao]{Li2022ELEVATERAB}
Chengkun Li, Haotian Liu, Liunian~Harold Li, Pengchuan Zhang, Jyoti Aneja,
  Jianwei Yang, Ping Jin, Yong~Jae Lee, Houdong Hu, Zicheng Liu, and Jianfeng
  Gao.
\newblock Elevater: A benchmark and toolkit for evaluating language-augmented
  visual models.
\newblock \emph{ArXiv}, abs/2204.08790, 2022.

\bibitem[Maji et~al.(2013)Maji, Rahtu, Kannala, Blaschko, and
  Vedaldi]{Maji2013FineGrainedVC}
Subhransu Maji, Esa Rahtu, Juho Kannala, Matthew~B. Blaschko, and Andrea
  Vedaldi.
\newblock Fine-grained visual classification of aircraft.
\newblock \emph{ArXiv}, abs/1306.5151, 2013.

\bibitem[Nilsback \& Zisserman(2008)Nilsback and
  Zisserman]{Nilsback2008AutomatedFC}
Maria-Elena Nilsback and Andrew Zisserman.
\newblock Automated flower classification over a large number of classes.
\newblock \emph{Indian Conference on Computer Vision, Graphics \& Image
  Processing}, pp.\  722--729, 2008.

\bibitem[Parkhi et~al.(2012)Parkhi, Vedaldi, Zisserman, and
  Jawahar]{parkhi2012cats}
Omkar~M Parkhi, Andrea Vedaldi, Andrew Zisserman, and CV~Jawahar.
\newblock Cats and dogs.
\newblock In \emph{{IEEE Conference on Computer Vision and Pattern Recognition
  (CVPR)}}, pp.\  3498--3505, 2012.

\bibitem[Radford et~al.(2021)Radford, Kim, Hallacy, Ramesh, Goh, Agarwal,
  Sastry, Askell, Mishkin, Clark, Krueger, and
  Sutskever]{Radford2021LearningTV}
Alec Radford, Jong~Wook Kim, Chris Hallacy, Aditya Ramesh, Gabriel Goh,
  Sandhini Agarwal, Girish Sastry, Amanda Askell, Pamela Mishkin, Jack Clark,
  Gretchen Krueger, and Ilya Sutskever.
\newblock Learning transferable visual models from natural language
  supervision.
\newblock In \emph{{International Conference on Machine Learning (ICML)}},
  2021.

\bibitem[Recht et~al.(2019)Recht, Roelofs, Schmidt, and Shankar]{Recht2019DoIC}
Benjamin Recht, Rebecca Roelofs, Ludwig Schmidt, and Vaishaal Shankar.
\newblock Do imagenet classifiers generalize to imagenet?
\newblock In \emph{{International Conference on Machine Learning (ICML)}},
  2019.

\bibitem[Shazeer \& Stern(2018)Shazeer and Stern]{shazeer2018adafactor}
Noam Shazeer and Mitchell Stern.
\newblock Adafactor: Adaptive learning rates with sublinear memory cost.
\newblock In \emph{{International Conference on Machine Learning (ICML)}}, pp.\
   4596--4604. PMLR, 2018.

\bibitem[Shen et~al.(2019)Shen, Nguyen, Wu, Chen, Chen, Jia, Kannan, Sainath,
  Cao, Chiu, et~al.]{shen2019lingvo}
Jonathan Shen, Patrick Nguyen, Yonghui Wu, Zhifeng Chen, Mia~X Chen, Ye~Jia,
  Anjuli Kannan, Tara Sainath, Yuan Cao, Chung-Cheng Chiu, et~al.
\newblock Lingvo: a modular and scalable framework for sequence-to-sequence
  modeling.
\newblock \emph{ArXiv}, abs/1902.08295, 2019.

\bibitem[Singh et~al.(2022)Singh, Hu, Goswami, Couairon, Galuba, Rohrbach, and
  Kiela]{singh2022flava}
Amanpreet Singh, Ronghang Hu, Vedanuj Goswami, Guillaume Couairon, Wojciech
  Galuba, Marcus Rohrbach, and Douwe Kiela.
\newblock Flava: A foundational language and vision alignment model.
\newblock In \emph{{IEEE Conference on Computer Vision and Pattern Recognition
  (CVPR)}}, pp.\  15638--15650, 2022.

\bibitem[Touvron et~al.(2021)Touvron, Cord, Douze, Massa, Sablayrolles, and
  J'egou]{Touvron2021TrainingDI}
Hugo Touvron, Matthieu Cord, Matthijs Douze, Francisco Massa, Alexandre
  Sablayrolles, and Herv'e J'egou.
\newblock Training data-efficient image transformers \& distillation through
  attention.
\newblock In \emph{{International Conference on Machine Learning (ICML)}},
  2021.

\bibitem[Udandarao et~al.(2022)Udandarao, Gupta, and
  Albanie]{Udandarao2022SuSXTN}
Vishaal Udandarao, Ankush Gupta, and Samuel Albanie.
\newblock Sus-x: Training-free name-only transfer of vision-language models.
\newblock \emph{ArXiv}, abs/2211.16198, 2022.

\bibitem[Vinyals et~al.(2015)Vinyals, Toshev, Bengio, and
  Erhan]{vinyals2015show}
Oriol Vinyals, Alexander Toshev, Samy Bengio, and Dumitru Erhan.
\newblock Show and tell: A neural image caption generator.
\newblock In \emph{{IEEE Conference on Computer Vision and Pattern Recognition
  (CVPR)}}, pp.\  3156--3164, 2015.

\bibitem[Wang et~al.(2019)Wang, Ge, Xing, and Lipton]{Wang2019LearningRG}
Haohan Wang, Songwei Ge, Eric~P. Xing, and Zachary~Chase Lipton.
\newblock Learning robust global representations by penalizing local predictive
  power.
\newblock In \emph{{Neural Information Processing Systems (NeurIPS)}}, 2019.

\bibitem[Wang et~al.(2022)Wang, Yu, Yu, Dai, Tsvetkov, and
  Cao]{DBLP:conf/iclr/WangYYDT022}
Zirui Wang, Jiahui Yu, Adams~Wei Yu, Zihang Dai, Yulia Tsvetkov, and Yuan Cao.
\newblock Simvlm: Simple visual language model pretraining with weak
  supervision.
\newblock In \emph{{International Conference on Learning Representations
  (ICLR)}}. OpenReview.net, 2022.

\bibitem[Wortsman et~al.(2022)Wortsman, Ilharco, Kim, Li, Kornblith, Roelofs,
  Lopes, Hajishirzi, Farhadi, Namkoong, et~al.]{wortsman2022robust}
Mitchell Wortsman, Gabriel Ilharco, Jong~Wook Kim, Mike Li, Simon Kornblith,
  Rebecca Roelofs, Raphael~Gontijo Lopes, Hannaneh Hajishirzi, Ali Farhadi,
  Hongseok Namkoong, et~al.
\newblock Robust fine-tuning of zero-shot models.
\newblock In \emph{{IEEE Conference on Computer Vision and Pattern Recognition
  (CVPR)}}, pp.\  7959--7971, 2022.

\bibitem[Yu et~al.(2022)Yu, Wang, Vasudevan, Yeung, Seyedhosseini, and
  Wu]{Yu2022CoCaCC}
Jiahui Yu, Zirui Wang, Vijay Vasudevan, Legg Yeung, Mojtaba Seyedhosseini, and
  Yonghui Wu.
\newblock Coca: Contrastive captioners are image-text foundation models.
\newblock \emph{ArXiv}, abs/2205.01917, 2022.

\bibitem[Yuan et~al.(2021)Yuan, Chen, Chen, Codella, Dai, Gao, Hu, Huang, Li,
  Li, Liu, Liu, Liu, Lu, Shi, Wang, Wang, Xiao, Xiao, Yang, Zeng, Zhou, and
  Zhang]{Yuan2021FlorenceAN}
Lu~Yuan, Dongdong Chen, Yi-Ling Chen, Noel C.~F. Codella, Xiyang Dai, Jianfeng
  Gao, Houdong Hu, Xuedong Huang, Boxin Li, Chunyuan Li, Ce~Liu, Mengchen Liu,
  Zicheng Liu, Yumao Lu, Yu~Shi, Lijuan Wang, Jianfeng Wang, Bin Xiao, Zhen
  Xiao, Jianwei Yang, Michael Zeng, Luowei Zhou, and Pengchuan Zhang.
\newblock Florence: A new foundation model for computer vision.
\newblock \emph{ArXiv}, abs/2111.11432, 2021.

\bibitem[Zhai et~al.(2022)Zhai, Wang, Mustafa, Steiner, Keysers, Kolesnikov,
  and Beyer]{zhai2022lit}
Xiaohua Zhai, Xiao Wang, Basil Mustafa, Andreas Steiner, Daniel Keysers,
  Alexander Kolesnikov, and Lucas Beyer.
\newblock Lit: Zero-shot transfer with locked-image text tuning.
\newblock In \emph{{IEEE Conference on Computer Vision and Pattern Recognition
  (CVPR)}}, pp.\  18123--18133, 2022.

\bibitem[Zhang et~al.(2022)Zhang, Fang, Gao, Zhang, Li, Dai, Qiao, and
  Li]{zhang2021tip}
Renrui Zhang, Rongyao Fang, Peng Gao, Wei Zhang, Kunchang Li, Jifeng Dai,
  Yu~Qiao, and Hongsheng Li.
\newblock Tip-adapter: Training-free clip-adapter for better vision-language
  modeling.
\newblock In \emph{{European Conference on Computer Vision (ECCV)}}, 2022.

\bibitem[Zhou et~al.(2022{\natexlab{a}})Zhou, Yang, Loy, and
  Liu]{Zhou2022LearningTP}
Kaiyang Zhou, Jingkang Yang, Chen~Change Loy, and Ziwei Liu.
\newblock Learning to prompt for vision-language models.
\newblock \emph{{International Journal on Computer Vision (IJCV)}},
  130:\penalty0 2337--2348, 2022{\natexlab{a}}.

\bibitem[Zhou et~al.(2022{\natexlab{b}})Zhou, Yang, Loy, and
  Liu]{zhou2022cocoop}
Kaiyang Zhou, Jingkang Yang, Chen~Change Loy, and Ziwei Liu.
\newblock Conditional prompt learning for vision-language models.
\newblock In \emph{{IEEE Conference on Computer Vision and Pattern Recognition
  (CVPR)}}, 2022{\natexlab{b}}.

\bibitem[Zhu et~al.(2023)Zhu, Zhang, He, Zhou, Wang, Zhao, and
  Gao]{Zhu2023NotAF}
Xiangyang Zhu, Renrui Zhang, Bowei He, Aojun Zhou, Dong Wang, Bin Zhao, and
  Peng Gao.
\newblock Not all features matter: Enhancing few-shot clip with adaptive prior
  refinement.
\newblock \emph{ArXiv}, abs/2304.01195, 2023.

\end{thebibliography}
\bibliographystyle{iclr2023_conference}

\clearpage
\appendix

\section{Appendix}

\subsection{Optimization}

We use the Adafactor optimizer~\citep{shazeer2018adafactor} with $\beta_1=0.9$, $\beta_2=0.999$, and a weight decay ratio of 0.01. All input images are first rescaled to $580\times580$ and then randomly cropped to the size of $540\times540$. We further apply RandAugment~\citep{Cubuk2020RandaugmentPA} and label smoothing in our data preprocessing pipeline. Our model is implemented in the Lingvo framework using Tensorflow~\citep{shen2019lingvo}.

\subsection{Analysis of category name initialization}
\label{sec:analysis-cni}

In this section, we delve deeper into how the proposed category name initialization helps in few-shot learning with large vision-language models. Vision-language models are adept at zero-shot inference without knowing any class names from downstream tasks. However, the zero-shot performance heavily depends on the domain gap and data distribution, thus varying on different downstream tasks. By leveraging a few training examples from the target domain, the pretrained vision-language models can adapt to the target domain.

%
% Specifically, we validate its universality to several vision-language models,   by conducting different experiments for comparison.
\paragraph{Improvement upon zero-shot performance.} We first examine how category name initialization improves zero-shot performance. As illustrated in Table~\ref{tab:few-shot-imagenet-variants} and Table~\ref{tab:few-shot-benchmark}, category name initialization enhances performance across all datasets. The improvement in performance from zero-shot to five-shot varies depending on the dataset. For instance, CoCa-2B on ImageNet saw a 1.32\% increase in performance, whereas EuroSAT saw a 36.04\% increase. CoCa's impressive zero-shot performance on ImageNet is leaving less room for few-shot learning. Nonetheless, the performance gain achieved through our category name initialization is noteworthy, as some other methods may not achieve comparable improvements, which will be discussed below. We also contend that our few-shot performance is not solely attributable to the strong pretrained CoCa model but also to our proposed category name initialization. For example, CoCa-2B's zero-shot performance on EuroSAT is 49.74\%, which is lower than that of most other approaches. However, with our category name initialization, it achieves 85.78\%, outperforming other approaches in the five-shot setting.

%the category name initialization did help the Though the performance gain on ImageNet is relatively smaller   is because that the room for     be We first observe the relative improvement upon  Leveraging the category name initialization, the model achieves the performance at the initial step of tTo better understand how category name initialization helps the few-shot learning, we compare the few-shot performance of with zero-shot performance 

\begin{table}[htp]
  \centering
  \renewcommand*{\arraystretch}{1.1}
  \caption{Comparing other fine-tuning methods on ImageNet and its variants.  We use IN as the abbreviation for ImageNet, and CNI for category name initialization. The second column means how much training data per class is used for finetuning. 0 shot means the pretrained vision-language model is directly evaluated without finetuning. All the numbers under the last five columns denote the top-1 test accuracy. 
  }
  \label{tab:comparison-finetuning-imagenet-variants}
  \scalebox{1}{
    \begin{tabular}{lcccccc}
        \Xhline{1pt}
         Model & Shot & IN & IN-V2  & IN-R & IN-A & IN-Sketch \\
        \hline
        CoCa-base & 0 & 82.26 & 76.32 & 93.16 &  76.17  & 71.43 \\
        \hline
        \multirow{2}[0]{*}{CoCa-base+Linear Probing} & 1 & 57.49 & 54.20 & 69.19 & 53.38 & 47.94 \\
         & 5 & 79.33 & 73.18 & 90.02 & 73.18 & 68.03 \\
        \hline
        \multirow{2}[0]{*}{CoCa-base+Full Fintuning} & 1 & 43.77 & 41.64 & 55.98 & 40.31 & 33.29 \\
         & 5 & 60.90 & 54.32 & 71.20 & 54.34 & 49.25 \\
        \hline
        \multirow{2}[0]{*}{Coca-base+CoOp} & 1 & 79.85 & 73.21 & 89.88 & 76.42 & 65.81 \\
         & 5 &  81.01 & 75.81 & 92.58 & 76.55 & 71.27 \\
        \hline
        % \multirow{3}[0]{*}{CoCa-base (Ours)} & 0 & 82.26 & 76.62 & 93.16 &  76.17  & \\
         \multirow{2}[0]{*}{CoCa-base+CNI} & 1 & 82.35 & 76.47 & 93.37 & 77.00 & 71.61 \\
         & 5 & 83.47 &  77.23 & 93.22  &  77.23 &  71.35 \\
          % & 1\% ($\approx12.8$) & 83.91 & 75.57 & 92.21 & 71.95 & 69.73 \\
        \Xhline{1pt}
    \end{tabular}%
  }
\end{table}

\paragraph{Comparing with other fine-tuning methods.}
In order to further validate the efficacy of category name initialization, we compare it with several other finetuning methods. CoCa-base is chosen as the pretrained vision-language model and experiments are carried out on ImageNet with different finetuning methods, such as linear probing, full finetuning, CoOp~\citep{Zhou2022LearningTP}, and category name initialization. As demonstrated in Table~\ref{tab:comparison-finetuning-imagenet-variants}, all finetuning methods, except category name initialization, fail to improve over zero-shot CoCa when one or five training examples per class are used. Furthermore, full finetuning underperforms linear probing due to the fact that the number of training examples is not commensurate with the number of trainable parameters in few-shot learning. Although CoOp demonstrated better performance compared to linear probing and full finetuning, its one- or five-shot performance is slightly inferior to zero-shot CoCa. This suggests that CoCa's few-shot performance is not significantly improved by learning contextual prompts. On the other hand, category name initialization effectively improves the few-shot performance, which is challenging when the zero-shot performance of CoCa is significantly higher than that of other counterparts  such as CLIP~\citep{Radford2021LearningTV}, FLAVA~\citep{singh2022flava}, and so on.

\begin{figure}[htp]
    \centering
    \includegraphics[width=0.48\textwidth]{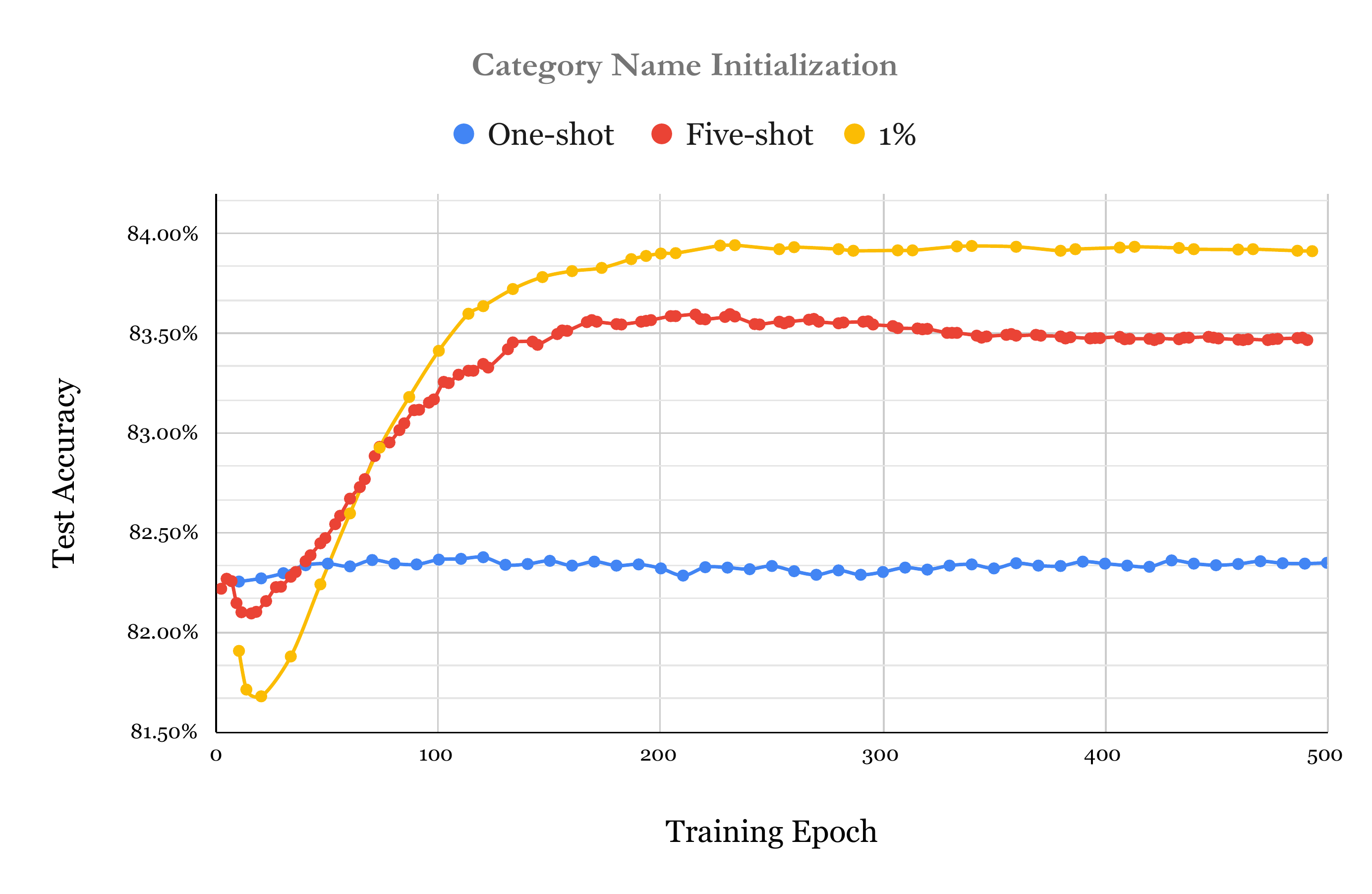}
    \hfil
    \includegraphics[width=0.48\textwidth]{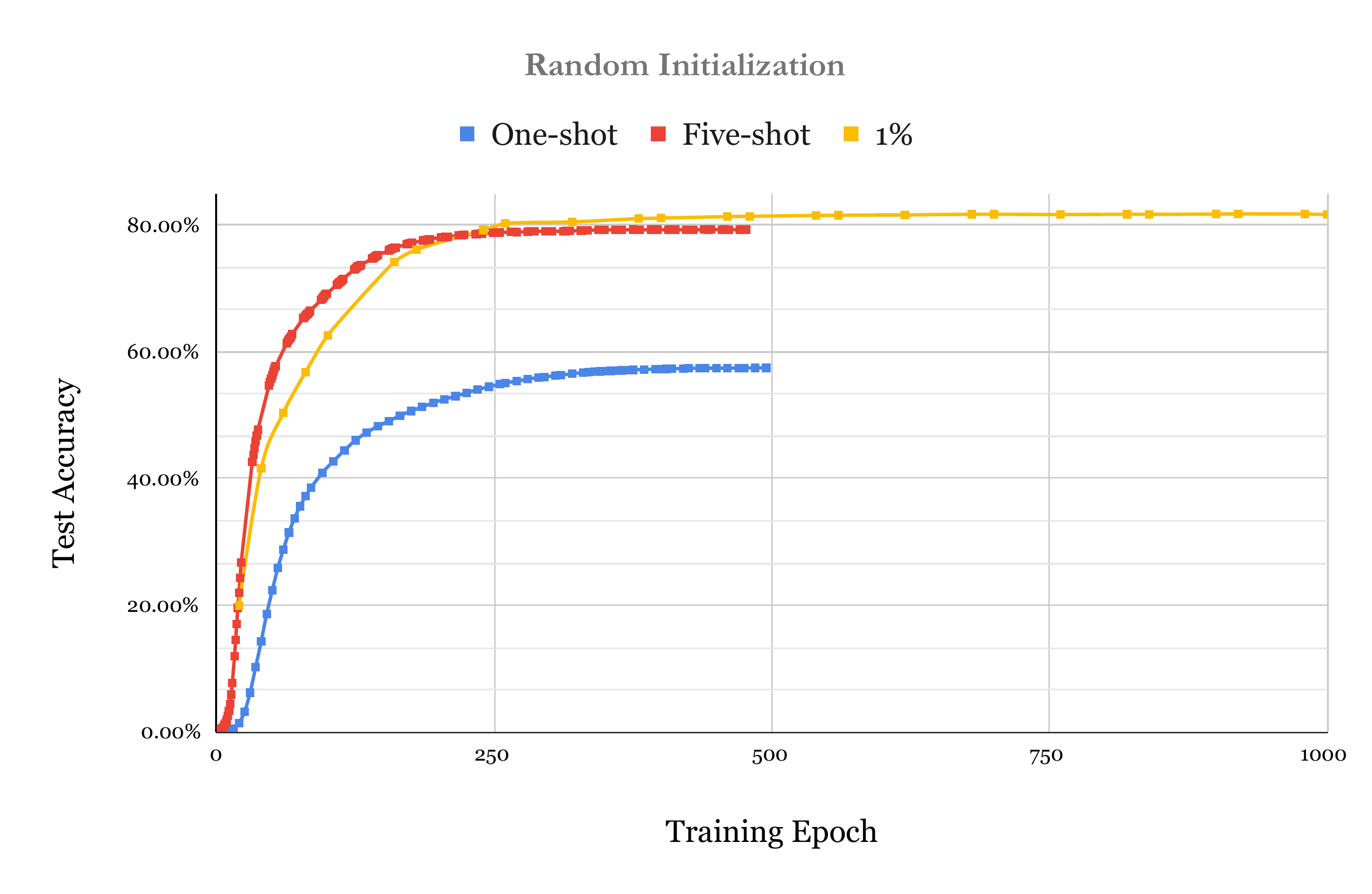}
    \caption{Comparison of test accuracy over the training epoch. We finetune the CoCa-base model with category name initialization or random initialization. Category name initialization provides better initial test accuracy and helps the model converge better and faster than random initialization. 
    % \llcao{the right figure is hard to read -- where shows it converges faster?}
    The test accuracy when using random initialization keeps increasing after 250 epochs, while the one using category name initialization converges around 200 epochs.
    }
    \label{fig:test_curve_comparison}
\end{figure}

\paragraph{Category name initialization vs random initialization.}
To gain a deeper understanding of the advantages of category name initialization, we compared it with random initialization. Figure~\ref{fig:test_curve_comparison} provides a more detailed comparison of the optimization process using the two initialization methods. By meticulously tuning the parameters, we set the initial learning rate to 1e-5 for category name initialization and 5e-5 for random initialization.
It is evident that category name initialization results in a better starting model with higher test accuracy compared to random initialization. Furthermore, the model utilizing category name initialization converges at a quicker pace than random initialization. This can be attributed to the fact that the test accuracy while using random initialization continues to increase even after 250 epochs, whereas the accuracy achieved with category name initialization plateaus around 200 epochs.
%We can observe that category name initialization provides a better starting model with higher test accuracy than random initialization. In addition, the model with category name initialization converges faster than random initialization. This is due to the observation that the test accuracy when using random initialization keeps increasing after 250 epochs, while the one using category name initialization converges around 200 epochs.
% However, both approach requires a large number of epochs to obtain the optimal performance. 

%after around 200 epochs, while test accuracy keeps going up when using random initialization (The right figure only shows 500 epochs for comparison).

\subsection{Comparing different initialization approaches}
\label{sec:different_initialization}
In the real world, we cannot guarantee that every classification task is associated with perfect category names. Sometimes we may only have digital labels like class ``1", ``2", ...; sometimes, the users may not speak fluent English. In these scenarios, we want to test how the model performs with different versions of category names.

Table~\ref{tab:different-init} compares the performance of using no category names (i.e., random initialization) with different variants of category names. The simplest situation is to use digits (class 1, 2, ...) as category names. This approach has little semantic information and thus provides no help with the few-shot performance. In contrast, category names in English and other languages significantly help the few-shot recognition. This is surprising because CoCa was trained in English-only text with limited knowledge of other languages.
However, thanks to sentence piece tokenizer~\citep{Kudo2018SentencePieceAS} and token sharing,
our method can still benefit from foreign language transfer to perform better than random initialization,
even though these foreign language names' performance is not as good as English names. 

\begin{table}[tp]
  \centering
  \caption{Comparison of category name initialization using digits or different languages. We use the same pretrained CoCa-base model for all category name initialization. The numbers below are \mbox{top-1} test accuracy on ImageNet.
  % \llcao{do you want to explain how you select the learning rate in the main text?}
  % \tx{I have explained in the ablation study - learning rates.}
  } 
  \label{tab:different-init}
  \scalebox{1}{
    \begin{tabular}{cccc}
        \Xhline{1pt}
        \specialcell{Category name\\ Initialization}  & Zero-shot & One-shot  & Five-shot \\ 
        \hline
        %Random	& 0.10	& 59.17	&  79.33 \\
        %No category name& 0.10	& 59.17	&  79.33 \\
        No  & N/A	& 59.17	&  79.33 \\
        Digits  & 0.10	 & 	53.60 & 	78.75 \\
        Korean  & 22.89 & 53.71 & 79.53 \\
        Russian	& 43.59	& 53.43	& 79.55 \\
        Germany & 29.24 & 63.15	& 79.90 \\
        Spanish & 	34.38 & 79.87 & 80.05 \\
        English & 82.26 & 82.35 &  83.58 \\
        \Xhline{1pt}
    \end{tabular}%
  }
\end{table}%

\begin{table}[tp]
  \centering
  \caption{Comparing the performance of using all category names or using 50\% of names (the other half will be initialized with random vectors) for initialization. The numbers below are top-1 test accuracy on ImageNet.}
  \label{tab:different-init-ratio}
  \scalebox{0.9}{
    \begin{tabular}{cccc}
        \Xhline{1pt}
        Initialization  & Zero-shot & One-shot  & Five-shot \\ 
        \hline
        %No category name& 0.10	& 59.17	&  79.33 \\
        No category name & N/A	& 59.17	&  79.33 \\
        50\% category names	 & 	44.36	 & 	66.82	 & 	80.67 \\
        100\% category names & 82.26 & 82.35 &  83.58 \\
        \Xhline{1pt}
    \end{tabular}%
  }
  \vspace{-2mm}
\end{table}%

Motivated by the above observation, we suspect only initialization with partial category information can still help. To verify this, we randomly choose 50\% of the category names as initialization while using random initialization for the rest. Table~\ref{tab:different-init-ratio} shows using 50\% of the names can still boost the one-shot accuracy from random initialization 59.17\% to 66.82\%, and five-shot accuracy from 79.33\% to 80.67\%. This suggests that our method is a promising tool when within-domain labels are not complete or come in different languages.

\subsection{When does the effect of category name initialization diminish? }

\begin{figure}[htp]
    \centering
    \subfloat[CoCa-base + CNI on ImageNet]{\includegraphics[width=0.5\textwidth]{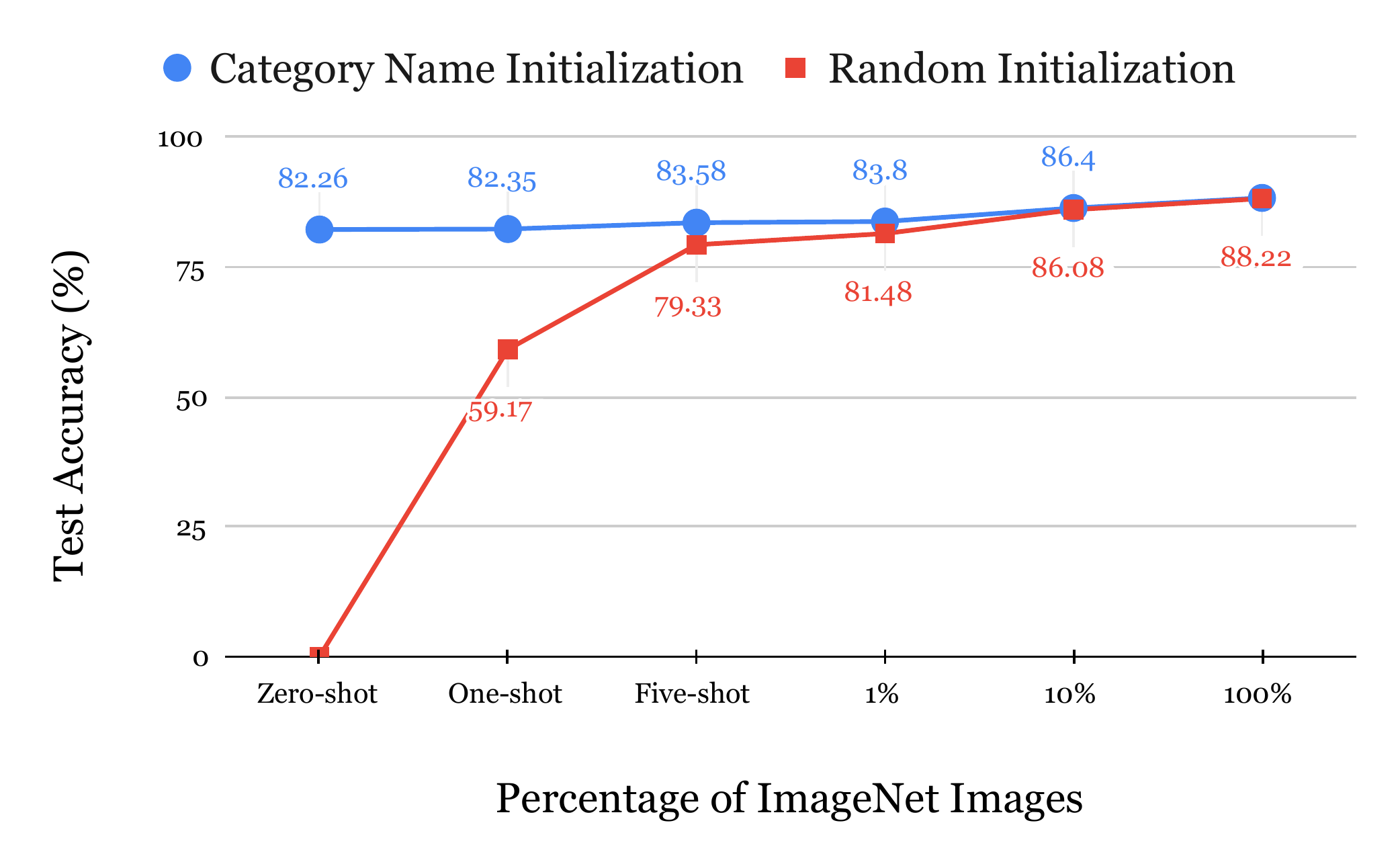}}
    \hfil
    \subfloat[CoCa-2B + CNI on Cifar100]{\includegraphics[width=0.5\textwidth]{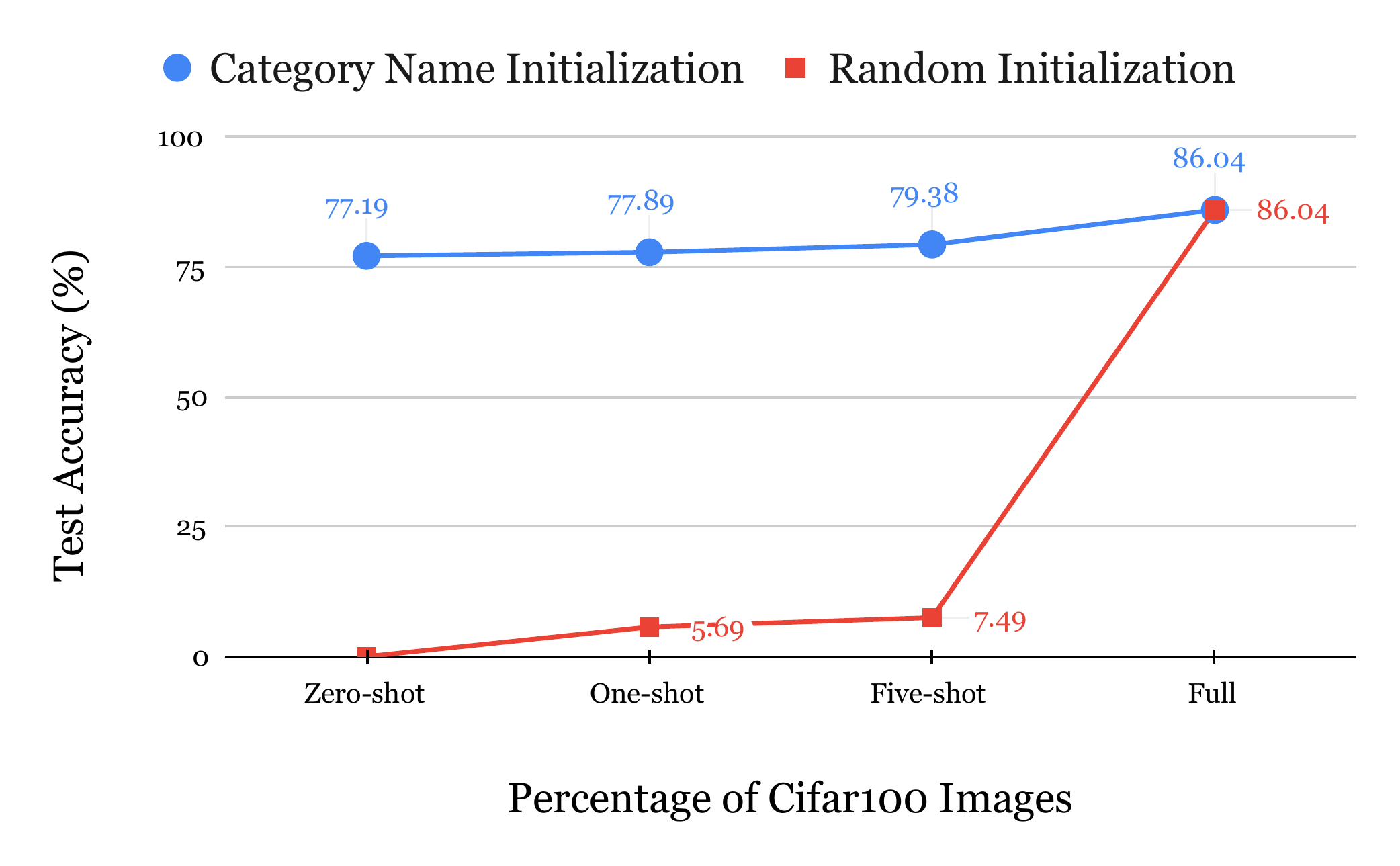}}
    \caption{Comparison of test accuracy over different percentages of training images. Category name initialization outperforms random initialization over different datasets, model architectures, and numbers of training data. }
    \label{fig:comparison_category_name_vs_random_init}
    \vspace{-2mm}
\end{figure}

To demonstrate the effectiveness of category name initialization, we set a baseline by using random initialization for comparison. We perform finetuning over different pretrained vision-language models using different numbers of training images. Specifically, we finetune CoCa-base on ImageNet and CoCa-2B on Cifar100. As shown in Figure~\ref{fig:comparison_category_name_vs_random_init}, category name initialization outperforms random initialization over different datasets, model architectures, and numbers of training data. The contribution of category name initialization diminishes as more training data is provided.

\subsection{Model distillation}

% To verify that  scalability of the category name initialization when scaling up large models,
% \llcao{I feel distillation is not related to category names initialization}, \tx{Without category name initialization, the performance would gets worse.}
% \tx{Two stories here}

First, to verify the scalability of CoCa, we carry out few-shot experiments using two different pretrained CoCa architectures: CoCa-base and CoCa-2B, under different numbers of training data. 
Abandoning the uni-modal and multi-modal text decoders, CoCa-base and CoCa-2B contain 96M and 1B parameters for downstream image classification tasks. 
As shown in Table~\ref{tab:few-shot-different-coca-imagenet-distillation}, we can observe the trend that bigger models do better and more shots help. 

\begin{table}[htp]
  \centering
  \caption{Few-shot results of different CoCa-models on ImageNet. 
  % \llcao{One table should be consistent, and tell one story. I feel you can remove CoCa-large here and focus on CoCa base and CoCa 2B. }
  }
  \label{tab:few-shot-different-coca-imagenet-distillation}
  \scalebox{0.92}{
    \begin{tabular}{rcccc}
        \Xhline{1pt}
         Model & Zero-shot & One-shot & Five-shot & 1\% \\
        \hline
        CoCa-2B & 86.19 & 86.15	& 87.37 & 87.90 \\
        CoCa-base & 82.26 & 82.35 & 83.58 & 83.80 \\
        + distillation & - & - & - & 84.81 \\
        % CoCa-large & 84.84  & & 85.98  & 86.51 \\
       \Xhline{1pt}
    \end{tabular}%
  }
\end{table}%

As large models give better performance, it is intuitive to consider knowledge distillation, i.e., taking the prediction of a teacher model as guidance to train a student model. We take the finetuned CoCa-2B using 1\% ImageNet images as the teacher model and CoCa-base as our student model. In addition to 1\% labeled ImageNet images, we also leverage other unlabeled images for knowledge distillation. The teacher model weights are frozen during the finetuning process, and the student model weights are updated according to two loss objectives. The first is the supervised loss, where we compute the cross entropy between the student model prediction and labels on the 1\% labeled ImageNet images. The other one is the distillation loss computed over all unlabeled data. In contrast to few-shot finetuning, where we finetune only the last few layers, we here finetune the whole student model because the distillation loss is computed over many unlabeled images. As shown in Table~\ref{tab:few-shot-different-coca-imagenet-distillation}, we have achieved a 1.01\% accuracy gain (from 83.80\% to 84.81\%) for CoCa-base by distilling from a bigger finetuned teacher model -- CoCa-2B.

\subsection{Ablation studies}
\label{sec:ablation-study}

In this section, we analyze several important factors that influence the few-shot performance. We use CoCa-base for our ablation study in the below.

\paragraph{Finetuning layers.}
We first use ImageNet~\citep{Deng2009ImageNetAL} to study the performance of the CoCa-base model by selecting different finetuning layers in several few-shot learning scenarios. We consider random initialization as the baseline for comparison. In our notation, P stands for the image pooler, and L stands for the linear projector.
For both category name initialization and random initialization, we try three different optimization strategies: 1) optimize L; 2) optimize P + L; 3) optimize All layers. 
Note that we have extensively tried various hyper-parameters (e.g., initial learning rate) and present the best number for each setting.

\begin{table}[htp]
  \centering
  \caption{Comparison of different finetuning layers for random initialization. 
  P: image pooler; L: linear projector; All: all layers. The best performance of each column is in {\bf bold}.
  }
  \label{tab:different-layers-random-init}
  \scalebox{1}{
    \begin{tabular}{ccccc}
        \Xhline{1pt}
        \specialcell{Finetuning\\ Layers} & One-shot & Five-shot & 1\% & 100\%  \\ 
        \hline
        L	& 49.38  & 69.64 &  76.53 & 85.62 \\
        P + L  &	{\bf 57.49} & {\bf 79.33} & {\bf 81.48} & {\bf 88.22} \\
        All  & 43.77 &	60.90 & 79.75 &  86.03 \\
        \Xhline{1pt}
    \end{tabular}%
  }
\end{table}%

\begin{table}[htp]
  \centering
  \caption{Comparison of different finetuning layers for category name initialization. P: image pooler; L: linear projector; All: all layers. The best performance of each column is in {\bf bold}.
  }
  \label{tab:different-layers-category-name-init}
  \scalebox{1}{
    \begin{tabular}{cccccc}
        \Xhline{1pt}
        \specialcell{Finetuning\\ Layers} & One-shot & Five-shot & 1\%  & 100\% \\  
        \hline
        L & 82.35 &  81.03  & 81.67 & 86.16 \\
        % C + L & 82.35 &	82.78 &	83.20 & 87.65 \\
        P + L & {\bf 82.35} & {\bf 83.58} &	{\bf 83.91}  & 88.25 \\
        All  &	82.28 & 82.63 &	83.63 & {\bf 88.35} \\
        \Xhline{1pt}
    \end{tabular}%
  }
\end{table}%

As shown in Table~\ref{tab:different-layers-random-init}, finetuning the image pooler and linear projector gives the best performance under all settings compared to the other two optimization strategies in random initialization.

To improve the performance of few-shot learning, we try category name initialization. 
In contrast to random initialization, we initialize the linear projector using the computed average text embeddings of the category names. 
Table~\ref{tab:different-layers-category-name-init} illustrates that the few-shot recognition performance is significantly improved. Besides, we can also observe that the finetuning P + L is the best optimization strategy for few-shot settings, while finetuning all layers works better given more training data.

\begin{figure}[htp]
    \centering
    \includegraphics[width=0.6\textwidth]{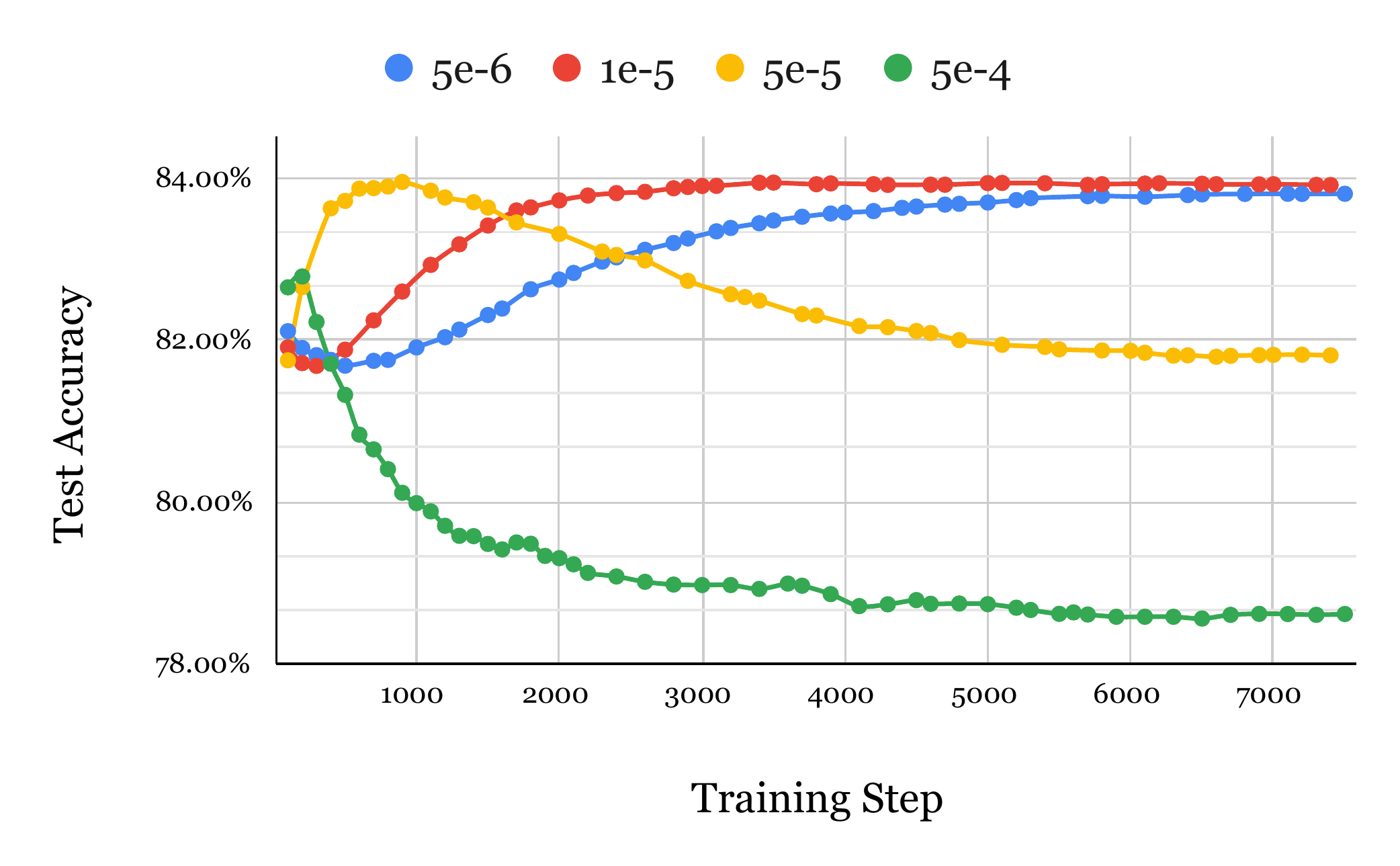}
    \caption{The top-1 test accuracy of finetuning CoCa-base on 1\% ImageNet using different initial learning rates.
    % \llcao{Optional: I was a bit confused with what what 53-6, 1-e-5 et al means}
    }
    \label{fig:learning_rates}
    % \vspace{-2mm}
\end{figure}

\paragraph{Learning rates.} We analyze the influence of the initial learning rate on few-shot learning. We set a batch size of 512, freeze the image encoder, and adopt a cosine learning rate schedule for the final three layers. Figure~\ref{fig:learning_rates} presents the top-1 test accuracy on ImageNet using different initial learning rates. Figure~\ref{fig:learning_rates} presents the top-1 test accuracy on ImageNet using different initial learning rates. A small initial learning rate (5e-6) results in a slow convergence rate, while a larger learning rate (5e-5) achieves faster convergence. However, despite reaching the highest test accuracy within 1000 training steps, the finetuning becomes unstable as the test accuracy declines right after the peak value. Conversely, using an even larger learning rate (5e-4) could prevent the surging phase, resulting in a downward trend of test accuracy. By contrast, selecting an appropriate learning rate (1e-5) is the key to stable and rapid few-shot finetuning. Unfortunately, there is no mathematical formula for determining the optimal initial learning rate since it varies across different datasets and depends on the batch size. We can adjust the initial learning rate by trial and observation, and these four test accuracy curves could indicate whether to enlarge or reduce the initial learning rate.

\begin{figure}[htp]
    \centering
    \includegraphics[width=0.6\textwidth]{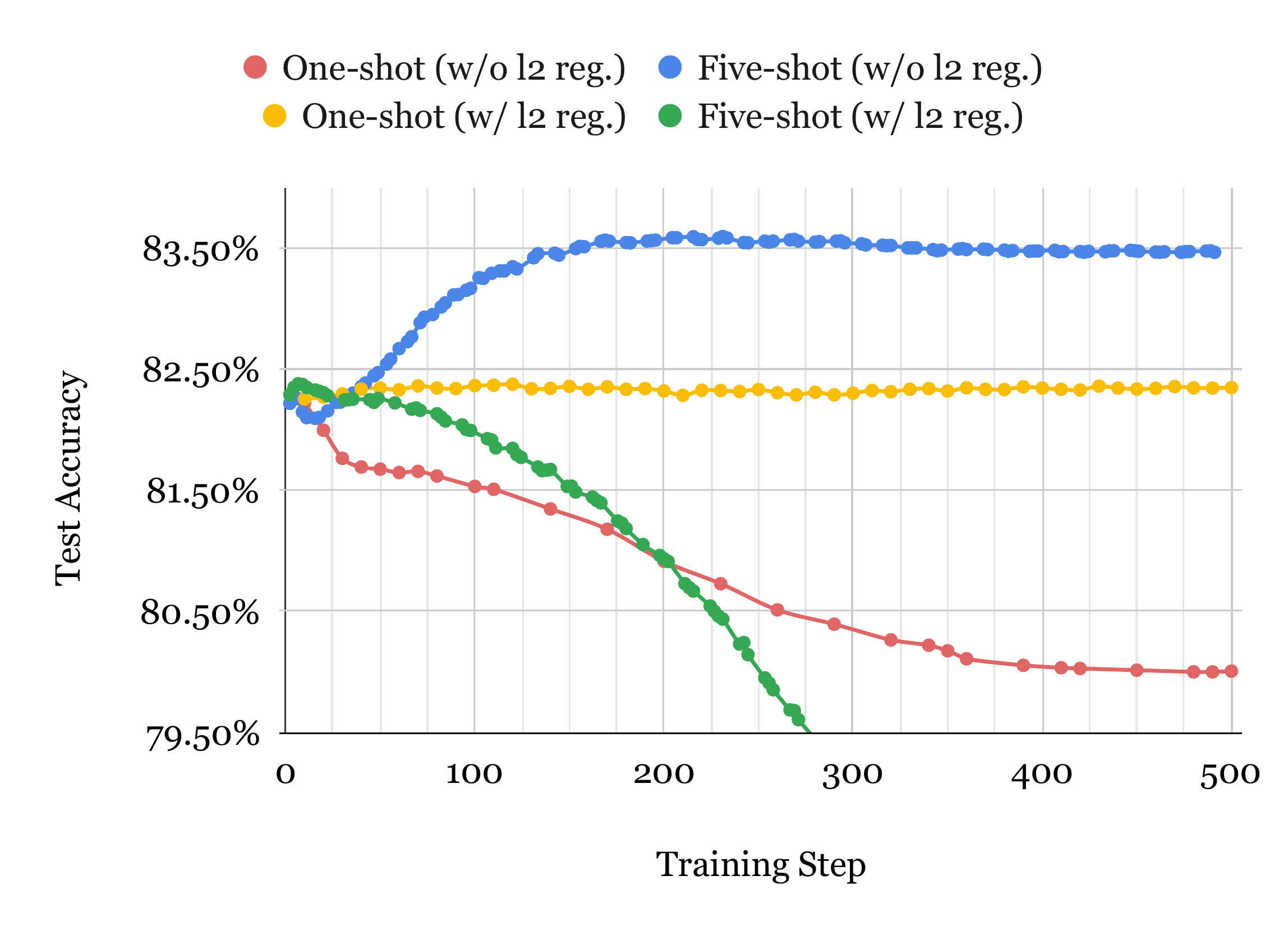}
    \caption{The effect of l2 weight regularization for one-shot and five-shot learning. We plot the top-1 test accuracy of CoCa-base on ImageNet vs. the training step. 
    % We have applied category name initialization  for all these four experiments.  
    The l2 weight regularization is beneficial to one-shot learning but harmful to five-shot learning.
    % \llcao{add: what is your conclusion/take-home-msg from this figure?}
    }
    \label{fig:l2_reg}
    \vspace{-2mm}
\end{figure}

\paragraph{L2 weight regularization.} Among all the few-shot settings, one-shot learning is the most special and intriguing one. Figure~\ref{fig:l2_reg} shows that the one-shot test accuracy (red) on ImageNet decreases even using category name initialization in finetuning in contrast to the five-shot one (blue). The pretrained feature gets distorted as the decision boundary could not be refined better using only one image per class. To address this issue, we leverage l2 weight regularization for one-shot learning. The test accuracy (yellow) increases stably from 82.26\% to 82.35\%. Though the performance gain is small, it is still nontrivial as the information provided by one-shot data is very limited to help a pretrained model. On the other hand, we find that applying the l2 weight regularization in five-shot learning could ruin the model adaptation, as depicted by the green curve. The reason is the l2 weight regularization, as an additional constraint, prevents the model from learning new knowledge from training data when the data contains relatively sufficient information for refining the decision boundary of the pretrained model.

\begin{figure}[htp]
    \centering
    \subfloat[Pretrained model]{\includegraphics[width=0.32\textwidth]{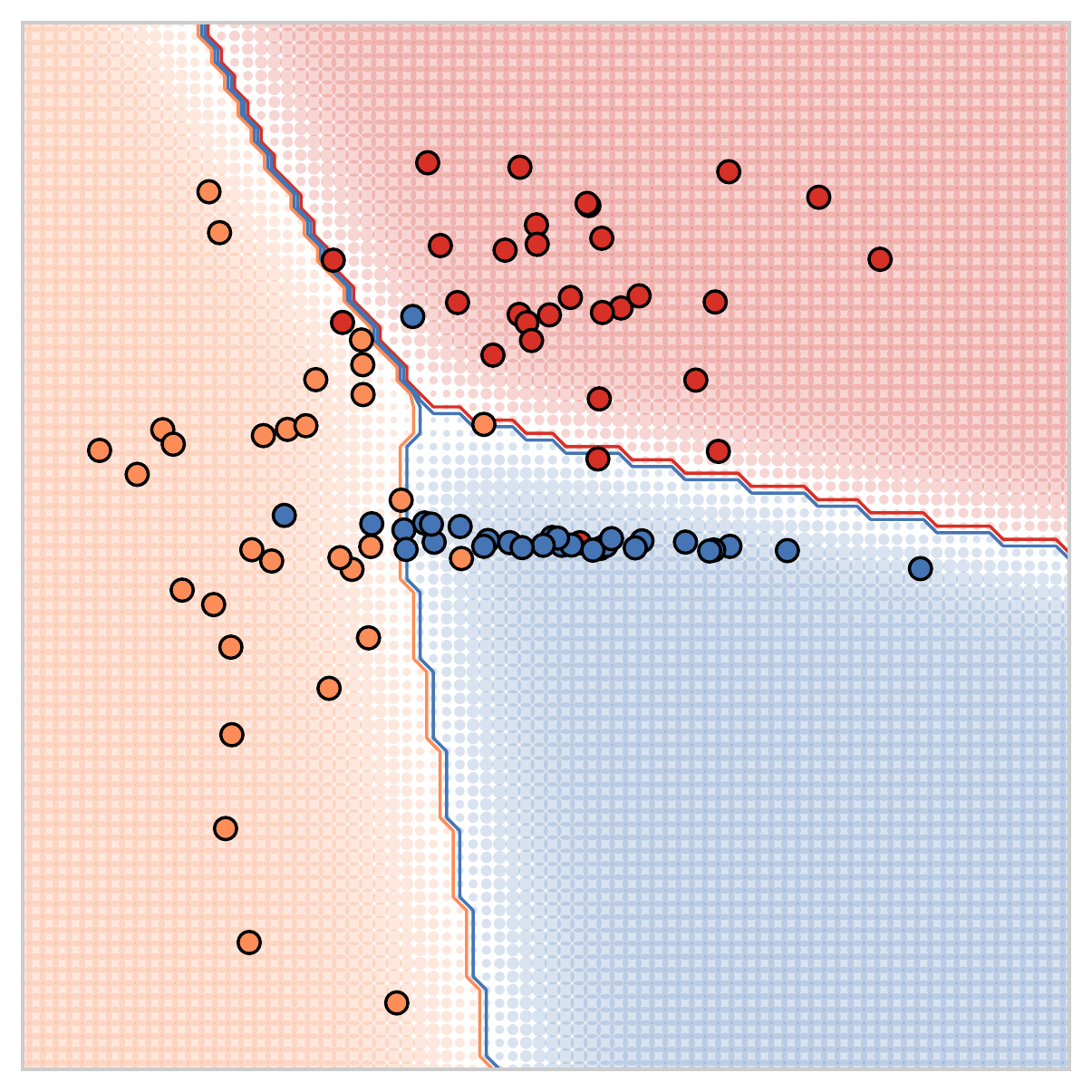}} \hfil
    \subfloat[No l2 regularization]{\includegraphics[width=0.32\textwidth]{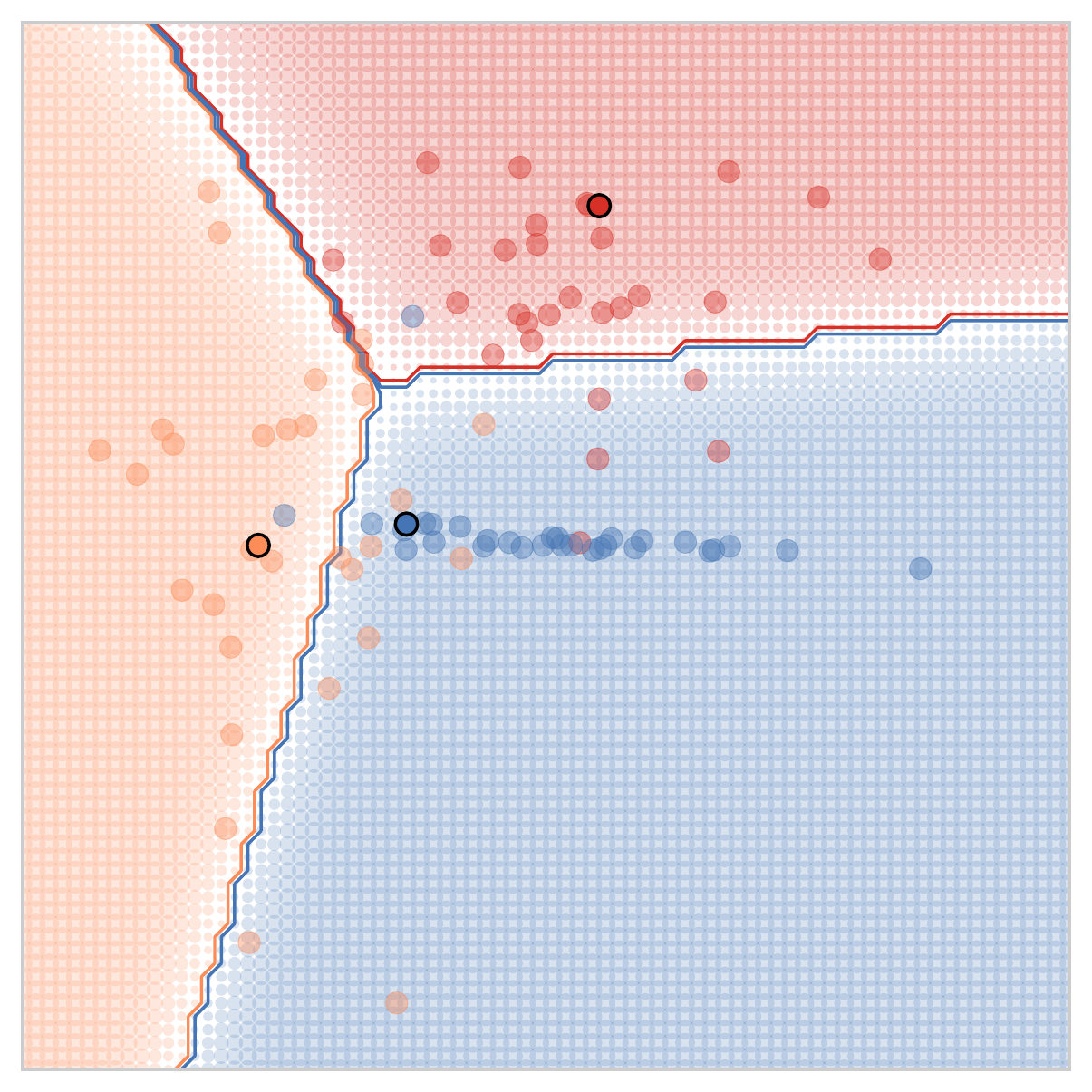}} \hfil
    \subfloat[L2 regularization]{\includegraphics[width=0.32\textwidth]{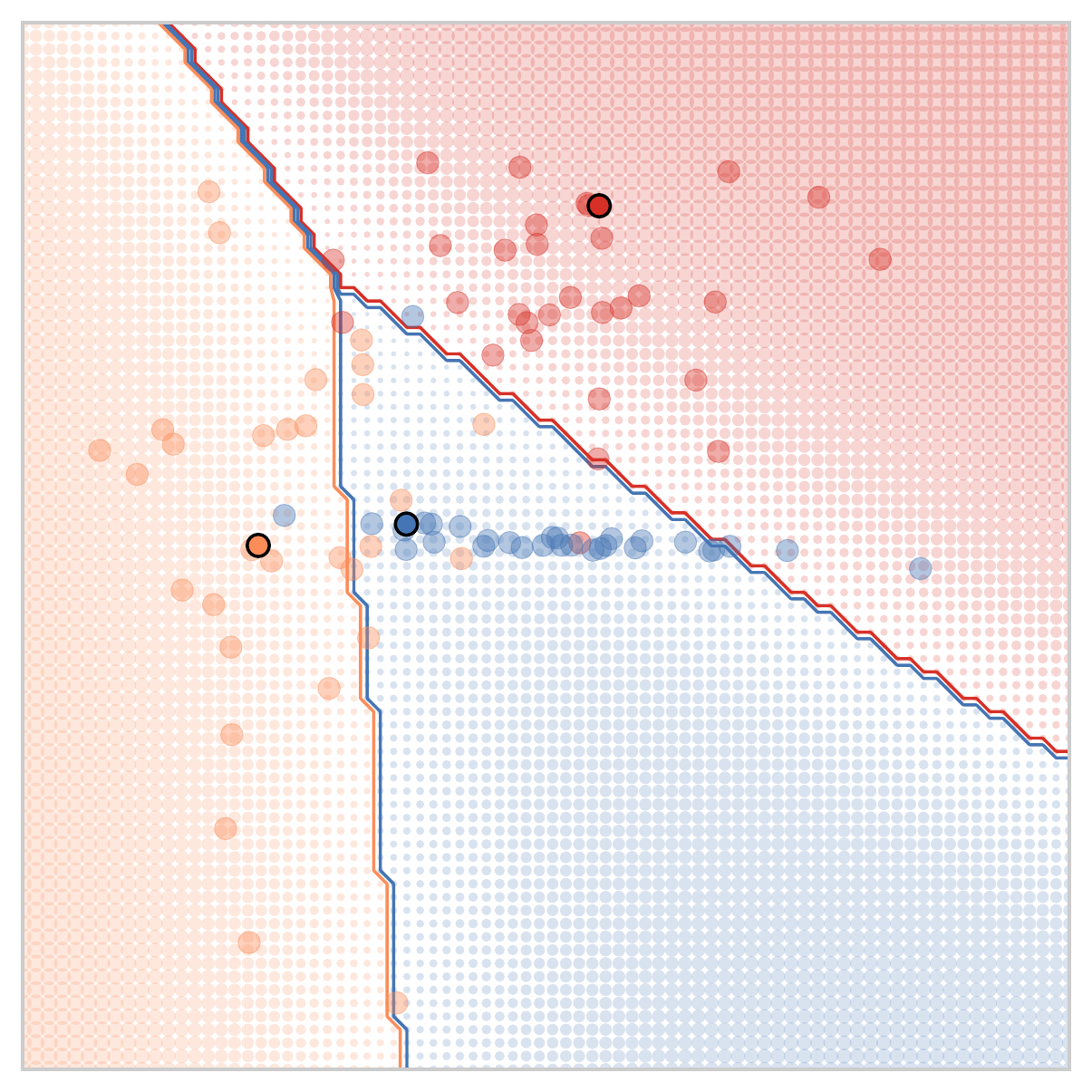}} 
    \caption{Visualization of decision boundary in one-shot learning. From left to right, The first subfigure displays the decision boundary of the pretrained model. In contrast, the second and third subfigures show the finetuned model without and with l2 weight regularization, respectively. Each model was trained using only one training example per class, with three classes retained for simplicity. The decision boundary does not shift significantly when finetuning on the one-shot dataset with l2 regularization. This indicates that the model's generalization ability is improved, as it is less likely to overfit to the training examples.
    }
    \label{fig:l2_reg-vis}
\end{figure}

\paragraph{L2 weight regularization.} Out of all the few-shot settings, one-shot learning is the most unique and intriguing. As illustrated in Figure~\ref{fig:l2_reg}, the one-shot test accuracy (in red) on ImageNet decreases even with category name initialization during finetuning, unlike the five-shot accuracy (in blue). Using only one training image per class can easily distort the decision boundary, as illustrated in Figure~\ref{fig:l2_reg-vis}.  We plot the decision boundary in Figure~\ref{fig:l2_reg-vis} for an illustration. Without l2 regularization, the decision boundary of the finetuned model is easily distorted by the limited training examples, resulting in a degradation from zero-shot performance. However, by applying l2 weight regularization for one-shot learning, the decision boundary does not deviate much compared with the decision boundary of the pretrained model. This is reflected in the steady increase of test accuracy from 82.26\% to 82.35\%, as depicted by the yellow curve in Figure~\ref{fig:l2_reg}.  Although the performance gain is small, it is still noteworthy since the information provided by one-shot data is limited in helping a pretrained model.  On the other hand, applying l2 weight regularization in five-shot learning could adversely affect the model adaptation, as shown by the green curve. The reason is that l2 weight regularization, acting as an additional constraint, restricts the model from learning new knowledge from the training data when sufficient information is available to refine the decision boundary of the pretrained model. It should be noted that all of the aforementioned phenomena are dependent on utilizing category name initialization. The decision boundary will lack discriminative power if category name initialization is not used. Therefore, adding l2 weight regularization would have no meaningful effect.

\end{document}